\documentclass[letterpaper]{article}
\usepackage[preprint]{aaai2027}
\usepackage[hyphens]{url}
\usepackage{graphicx}
\usepackage{natbib}
\usepackage{caption}
\usepackage{booktabs}
\usepackage{multirow}
\usepackage{array}
\usepackage{tabularx}
\usepackage{xcolor}
\usepackage[most]{tcolorbox}
\usepackage{pifont}
\usepackage{enumitem}
\usepackage{amsmath,amssymb}
\usepackage{algorithm}
\usepackage{algorithmic}
\usepackage{placeins}


\newcommand{\methodname}{\textsc{ContraMem}}
\newcommand{\method}{\methodname}
\newcommand{\gaia}{\textsc{GAIA2}}
\newcommand{\are}{\textsc{ARE}}

\newcommand{\reflector}{Reflector}
\newcommand{\curator}{Curator}

\newenvironment{fullwidthplate}{\begin{figure*}[t]\centering}{\end{figure*}}
\providecommand{\memorycardsfigurewidth}{0.94\textwidth}
\tcbset{
  bwplate/.style={
    enhanced, breakable,
    colback=white, colframe=black!70,
    colbacktitle=black!8, coltitle=black,
    fonttitle=\bfseries\small,
    boxrule=0.55pt, arc=0pt,
    left=6pt, right=6pt, top=5pt, bottom=5pt
  }
}
\newcommand{\promptimagepage}[2]{%
  \begin{figure*}[p]
    \centering
    \includegraphics[page=#2,width=0.985\textwidth]{#1}
  \end{figure*}%
}
\newcommand{\promptsubsection}[1]{%
  \refstepcounter{subsection}\label{#1}%
}
\newcolumntype{K}{>{\ttfamily\raggedright\arraybackslash}p{0.24\textwidth}}
\newcolumntype{V}{>{\raggedright\arraybackslash}X}

\title{CONTRAMEM: Learning Self-Evolving Procedural Memory from Contrasting Multi-Model Trajectories}
\author{%
Zheyuan Deng\textsuperscript{1}\equalcontrib, Binghang Lu\textsuperscript{2}\equalcontrib, Hanqi Feng\textsuperscript{3}, Shirley Huang\textsuperscript{4}, Dianzhuo Wang\textsuperscript{4}\\
Yuanda Xu\textsuperscript{5}, Zhiwei Zhang\textsuperscript{6}, Yige Sun\textsuperscript{7}, Changhong Mou\textsuperscript{8}, Runyu Zhang\textsuperscript{9}\\
Yuexing Hao\textsuperscript{9}, Barnabas Poczos\textsuperscript{3}, Guang Lin\textsuperscript{2}\corresponding, Xiaomin Li\textsuperscript{4}\corresponding
}
\affiliations{%
\textsuperscript{1}Brown University\quad
\textsuperscript{2}Purdue University\quad
\textsuperscript{3}Carnegie Mellon University\quad
\textsuperscript{4}Harvard University\\
\textsuperscript{5}Princeton University\quad
\textsuperscript{6}Pennsylvania State University\quad
\textsuperscript{7}Independent Researcher\\
\textsuperscript{8}Utah State University\quad
\textsuperscript{9}Massachusetts Institute of Technology
}

\begin{document}
\maketitle

\begin{abstract}
Autonomous computer-use agents are increasingly applied to long-horizon tasks
requiring coordinated application calls, persistent state tracking, and
verifier-sensitive writes, yet they remain prone to procedural
failures---misreading application state, tool semantics, or task progress.
Procedural memory promises more consistent decisions and less redundant
exploration, but constructing high-quality memory without model training
remains challenging. We introduce \methodname{}, a source-flexible, training-free framework for
self-evolving procedural memory that treats same-task outcome variation as
supervision: differences in correctness, efficiency, recovery, and failure
modes expose outcome-relevant procedural distinctions, distilled into a
compact bank of app-level Function Cards and task-level Skill Cards that
evolves through localized curation rather than append-only accumulation or
whole-bank rewriting. On held-out \gaia{}/\are{} computer-use tasks, \methodname{} more than
doubles the success rate across the three source-model targets (26.2\% to 55.3\%),
with consistent per-model gains (GPT-5.5: $27.5\!\to\!61.0$; Claude Sonnet
4.6: $28.0\!\to\!52.5$; DeepSeek V4 Pro: $23.0\!\to\!52.5$). The same bank
transfers unchanged to the unseen Qwen3.7 Plus ($18.5\!\to\!35.5$),
indicating transferable procedural knowledge rather than model-specific
behavior. The same construction carries over unchanged to AppWorld, beating both no
memory and its own single-source self-memory variant for all three mid-tier
agents on both public test splits. Under a matched trajectory budget, heterogeneous multi-model trajectories
yield stronger memory than self- or same-model multi-rollout memory: the
margin comes from contrastive behavioral diversity, not stronger source
agents or more sampling.

\end{abstract}

\section{Introduction}
\label{sec:introduction}

Large language models are increasingly deployed as autonomous computer-use
agents that interleave reasoning with action~\citep{yao2022react}, invoke application functions and external
tools~\citep{schick2023toolformer,patil2024gorilla,li2023api}, and issue
state-changing writes across realistic, multi-application
environments~\citep{lu2024toolsandbox,trivedi2024appworld,froger2026gaia2}.
A successful agent in this regime must select the correct function, construct
valid arguments, interpret asynchronous and often empty observations, maintain
pending obligations, and stop only once the task is genuinely complete---and
this difficulty does not disappear with scale: even strong models remain
brittle on long-horizon tasks with state dependencies, temporal constraints,
and verifier-sensitive
writes~\citep{trivedi2024appworld,froger2026gaia2}. A natural, training-free way to close this gap is procedural memory:
retained knowledge lets an agent avoid re-deriving equivalent searches,
repeating the same function misuse, or relapsing into identical recovery
failures~\citep{shinn2023reflexion,zhao2023expel,wang2023augmenting,park2023generativeagents,packer2023memgpt}.

\begin{figure*}[t]
  \centering
  \includegraphics[width=0.99\textwidth]{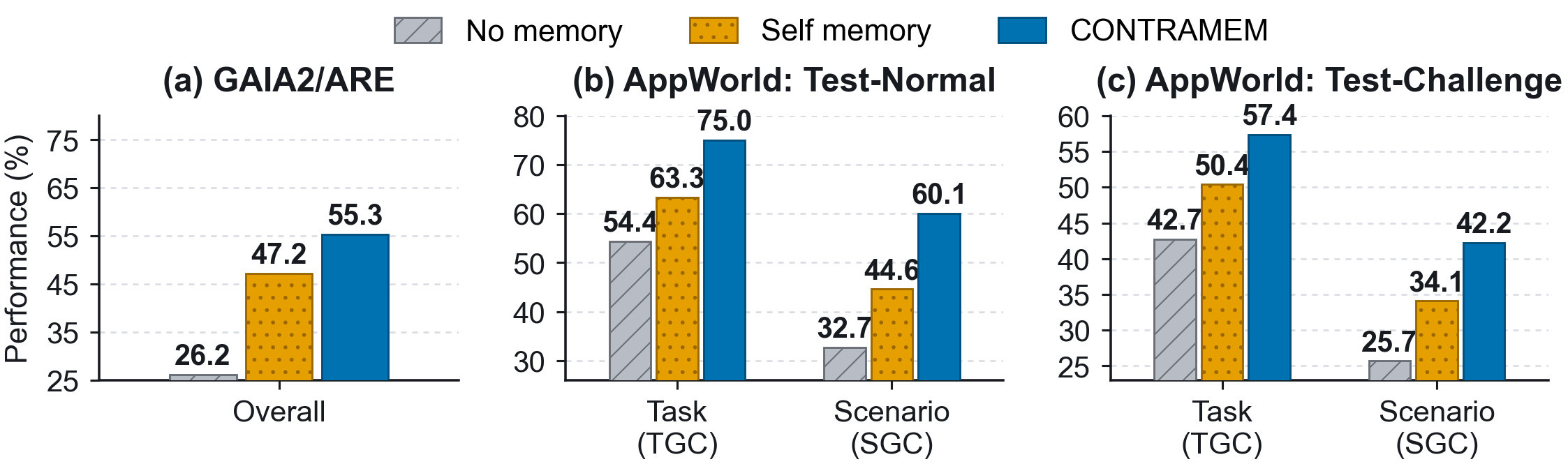}
  \caption{Held-out success averaged over three source targets (GAIA2/ARE)
  and target-macro TGC/SGC (AppWorld): \methodname{} beats self memory on
  every metric.}
  \label{fig:heldout_overall_transfer}
\end{figure*}

Yet turning past experience into high-quality procedural memory without
updating model weights is structurally difficult: raw retrieval transfers
brittle surface detail instead of reusable logic, per-trajectory summaries
miss the causal boundary separating success from failure, and same-model
resampling suffers from \emph{blind-spot inheritance}---a memory distilled from
one policy reproduces the very gaps that caused its failures. Append-only
accumulation compounds this with redundancy and drift: agents over-follow
retrieved experience, propagate stale errors, and degrade as continuously
updated memories accumulate
faults~\citep{fang2026trajectorymemory,xiong2025memorymanagement,zhang2026faultymemory,xu2025mem,zhang2025ace}.
The question is how to keep the bank compact, grounded, actionable, and
retrievable as it evolves.

We identify heterogeneous exploration as an underused source of such
memory---much as a team post-mortem that compares several attempts at the
same incident teaches more than any single log. When distinct models attempt
the same computer-use task, their behavior diverges informatively: one run
succeeds where another fails, one path is markedly shorter, one model
recovers where another stalls, or several fail at a common operation: same-task differences in
correctness, efficiency, recovery, and failure that are not noise to be
averaged away but supervision that multi-model practice already produces, at
no extra labeling cost. Examining multiple paths
surfaces stronger strategies than any single
execution~\citep{yao2023bot,besta2023got,guo2025deepseekr1,lin2025seagent,chang2026memcollab,tang2025agentkb};
for computer-use agents, such paths need not be consumed at test time and
discarded. Instead, they can be distilled offline into reusable memory for future
single-agent executions.

\begin{figure*}[t]
  \centering
  \includegraphics[width=0.99\textwidth]{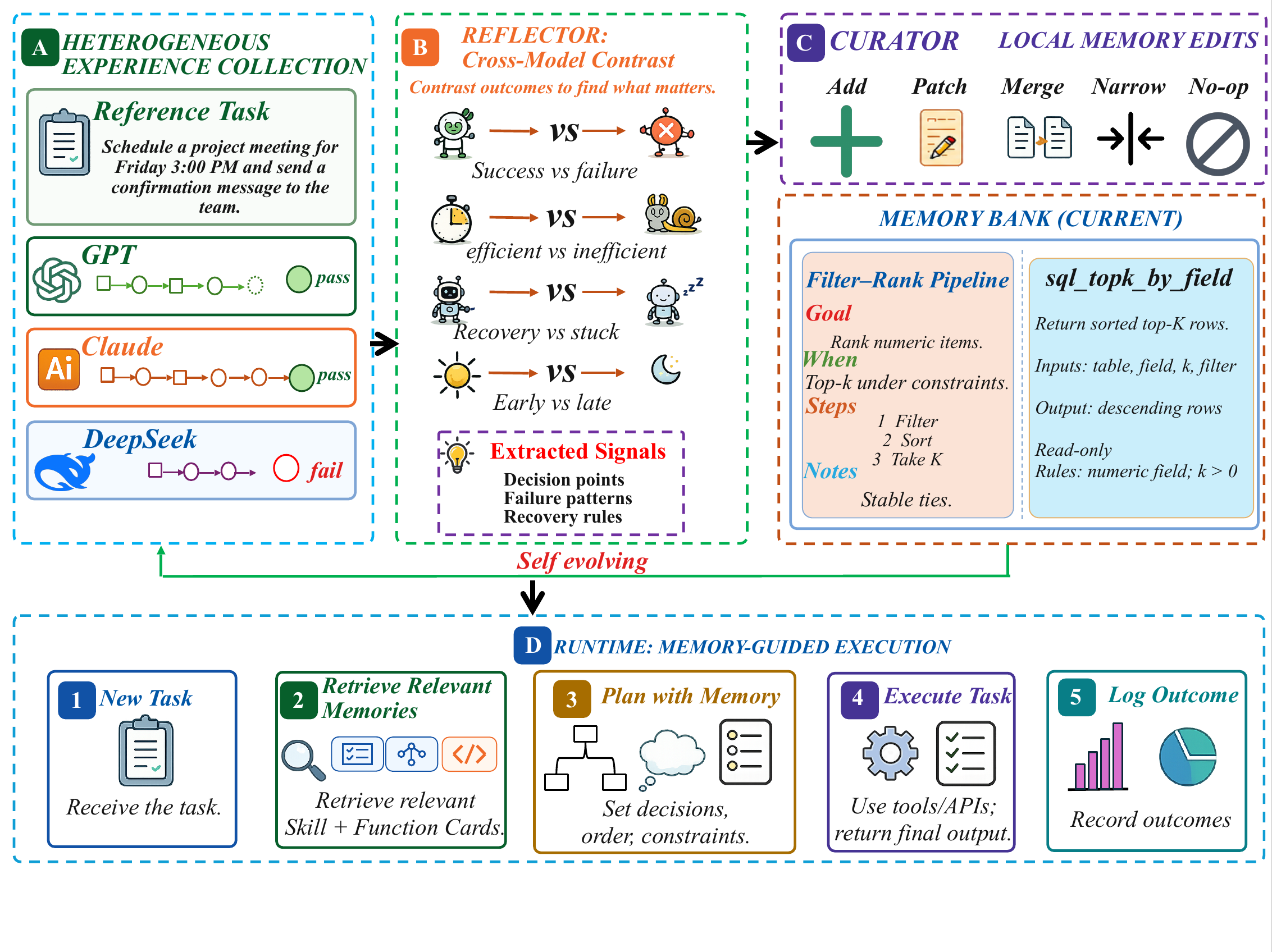}
  \caption{Overview of \methodname{}: heterogeneous same-task trajectories
(A) are compared by the \reflector{} to isolate outcome-relevant procedural
differences (B); the \curator{} uses these signals to refine Function and Skill
Cards in the current memory bank (C), from which a compact task-relevant subset
guides a single target agent at runtime (D).}
  \label{fig:contramem_overview}
\end{figure*}

Building on this view, we introduce \methodname{}, a training-free,
source-flexible framework that converts execution trajectories into a compact
procedural memory bank. Given one or more trajectories for a reference task,
\methodname{} extracts outcome-relevant distinctions along four
axes---correctness, efficiency, recovery, and recurring failure---and
distills them, rather than the raw traces, into \emph{Function Cards}
(app-level tool contracts) and \emph{Skill Cards} (task-level decision
rules), maintained by a local \curator{} through small, targeted edits.
Figure~\ref{fig:contramem_overview} summarizes this offline construction
and the subsequent frozen-bank runtime. \methodname{} does not require source diversity; it improves execution even from a single model's own trajectories. It reaches its strongest form, however, when sources are heterogeneous, transposing the central intuition of contrastive representation learning~\citep{oord2018cpc,chen2020simclr,he2020moco} from
representations to agent behavior: several models' attempts at one task are
multiple views of its procedural structure, and their disagreements form a
supervision signal that no single model can reliably supply, because
same-model rollouts inherit one policy and one set of blind spots. The analogy is conceptual, not algorithmic: no embeddings, no contrastive
objective. Multi-model exploration stays offline; deployment uses one target agent,
and no parameters are updated.

We evaluate \methodname{} on \gaia{}/\are{}, a dynamic app-agent benchmark
with asynchronous state changes, ambiguity, time-sensitive obligations, and
action-level write verification~\citep{froger2026gaia2,are}, with GPT-5.5,
Claude Sonnet 4.6, and DeepSeek V4 Pro as sources---a high-capability frontier
where gains reflect headroom beyond raw capability. Across the three source
target models, \methodname{} more than doubles held-out success
(26.2\%$\to$55.3\%), transfers the bank unchanged to Qwen3.7 Plus, an unseen
target ($18.5\!\to\!35.5$), and lands 30.0 points above the strongest prior memory system
(Figure~\ref{fig:memory_baselines}); Figure~\ref{fig:heldout_overall_transfer}
summarizes this progression alongside AppWorld, where the same construction
beats both no memory and target-specific self memory for all three mid-tier
agents on both public test splits~\citep{trivedi2024appworld}. Controlled
ablations confirm the gains stem from cross-model heterogeneity rather than
stronger sources or more sampling (Table~\ref{tab:components}).

Our contributions are threefold:
\begin{enumerate}
    \item \methodname{}: a source-flexible, training-free framework for
    self-evolving procedural memory, treating cross-model behavioral contrast
    as procedural supervision.
    \item A granularity-aware representation pairing app-level Function Cards
    with task-level Skill Cards, maintained by a localized \curator{}
    through targeted edits.
    \item Controlled evaluations that separately quantify the architecture's
    single-source gains and the added value of heterogeneous contrast, with
    transfer to an unseen target and cross-benchmark validation on
    AppWorld.
\end{enumerate}

\section{Related Work}
\label{sec:related_work}

\paragraph{Computer-use agents and app-agent benchmarks.}
Large language models have moved from text-only reasoning toward agents that
act: ReAct introduced a general reasoning--acting loop, while Toolformer,
Gorilla, and API-Bank studied how language models select and invoke external
tools and APIs
\citep{yao2022react,schick2023toolformer,patil2024gorilla,li2023api}.
Recent benchmarks stress long-horizon, stateful app-agent
behavior---conversational tool use in ToolSandbox, cross-application coding
agents in AppWorld, and dynamic asynchronous tasks with ambiguity, time
obligations, and action-level write verification in \gaia{}/\are{}
\citep{lu2024toolsandbox,trivedi2024appworld,froger2026gaia2}---exposing a gap \methodname{} targets.

\paragraph{Trajectory-derived and experience-replay memory.}
A large body of work converts past executions into reusable experience:
verbal reflections (Reflexion), lessons (ExpeL), executable skills (Voyager),
induced workflows (Agent Workflow Memory)
\citep{shinn2023reflexion,zhao2023expel,wang2023voyager,wang2024awm}, and
distilled thought templates or procedural memories
\citep{yang2024buffer,ouyang2026reasoningbank,cao2026reme}. Closer to our setting, Trajectory-Informed Memory Generation produces
strategy and recovery tips for future AppWorld tasks
\citep{fang2026trajectorymemory}, and Contextual Experience Replay keeps a
dynamic buffer of environment dynamics for web agents \citep{liu2025cer}. These methods, however, extract memory from individual trajectories or a single agent's experience, missing the contrastive boundary that explains why one execution succeeds while another fails. They also rarely separate function-level knowledge from task-level logic.
\methodname{} targets both gaps; its self-memory control instantiates this
single-agent line inside the same architecture, isolating exactly what
heterogeneous contrast adds.

\paragraph{Evolving contexts and memory management.}
A parallel line emphasizes that memory quality depends on maintenance~\citep{lu2026muon}. A-MEM
organizes agent memories as structured, linked notes that evolve as new
memories arrive \citep{xu2025mem}; ACE treats context as an evolving playbook
and argues that whole-context rewriting causes brevity bias and context
collapse, motivating incremental generation--reflection--curation updates
\citep{zhang2025ace}; and empirical studies confirm these failure modes in continuously updated
memories \citep{xiong2025memorymanagement,zhang2026faultymemory}. In its released
offline path, ACE builds a single playbook from an agent's own executions and
injects it in full; \methodname{} instead derives typed cards from same-task
cross-model contrasts, refines them through localized curation, and retrieves
only task-relevant guidance.

\paragraph{Cross-agent, cross-model, and multi-path learning.}
A final line studies how multiple solution paths or agents improve reasoning
and reuse: Tree and Graph of Thoughts explore multiple reasoning paths
\citep{yao2023bot,besta2023got}, SE-Agent evolves trajectories by revision
and recombination \citep{lin2025seagent}, and Agent KB aggregates
cross-framework trajectories \citep{tang2025agentkb}. Most directly related is MemCollab, which likewise contrasts trajectories
from different models on the same task \citep{chang2026memcollab}---but
toward transferability, distilling agent-invariant constraints so one memory
can be shared across backbones. \methodname{} uses the same disagreement for
a different end: rather than normalizing model differences away, it exploits them to raise
memory quality among frontier sources under a matched budget.

\section{Methodology}
\label{sec:method}

\subsection{Problem Setup}
\label{sec:method_setup}

\methodname{} constructs a typed memory bank from reference-task
trajectories and retrieves a small task-relevant subset for held-out
execution.

With $\mathcal{A}$ the ability set, tasks in each $a \in \mathcal{A}$ are
split into a reference set $\mathcal{D}^{a}_{\mathrm{ref}}$ and a held-out
set $\mathcal{D}^{a}_{\mathrm{test}}$; each source agent $m \in \mathcal{M}$
attempts each reference task $x_i$, producing a trajectory
$\tau_{i,m} = (e_1, \ldots, e_{T_{i,m}}, y_{i,m})$ of interaction events with
a verifier-assigned outcome. From these, \methodname{} constructs a bank
$\mathcal{B} = (\mathcal{F}, \{\mathcal{S}^{a}\})$: a global Function Card
bank $\mathcal{F}$ shared across abilities and ability-specific Skill Card
banks $\mathcal{S}^{a}$. At evaluation time, a target agent $g$, possibly
$g \notin \mathcal{M}$, receives guidance rendered by the deterministic
retriever $\rho(x,\mathcal{B})$ while attempting
$\mathcal{D}^{a}_{\mathrm{test}}$; no model parameters are updated.
Algorithms~1--2 in Appendix~A.8 summarize the complete construction and
runtime pipeline.

This separates two transfers: Function memory is global because a function
keeps its semantics across task families; Skill memory stays ability-specific
because one surface pattern implies different obligations---a message write
may be a final answer in search, a side effect in execution, or a
clarification boundary in ambiguity.

\paragraph{Trajectory normalization.}
Raw trajectories interleave model messages, app calls and observations,
environment notifications, verifier feedback, and final responses. We map each trace to a common event sequence recording source, type, and
position and, for app calls, the function, normalized arguments, and a
compact observation summary. Agent-callable functions ground Function Cards; environment and verifier
events are never tools but stay available for Skill reflection. Before any artifact reaches a construction model, task-specific values are
replaced under the placeholder policy (Appendices~B.3--B.5) while dates,
times, and operation order are retained.

Two intermediate artifacts derive from this sequence: a \emph{function
observation} captures one callable function's arguments, return form, side
effects, and trajectory outcome, while a \emph{task contrast packet} groups
one task's source trajectories---task text, outcomes, ordered action spans,
writes, final responses, and normalized verifier feedback---exposing where runs diverge while preserving the read/write/wait/finalize
order that determined each outcome.

\subsection{Function Card Construction}
\label{sec:function_cards}

Each Function Card compactly describes one callable app function; because a
function's interface and side effects are shared across tasks, its card is
reused across abilities. For every observed function $u$, a builder condenses its calls from all
reference trajectories (evidence about accepted arguments, return forms,
empty or error cases, and state-changing effects) into one card $f_u$: what
it does, how to call it safely, what it returns, and which mistakes need
caution. Cards remain strictly tool-level: every claim must be supported by an
observed call, and no card may contain a task plan, trajectory summary, model
identity, or verifier judgment. This keeps them useful both at task start and
immediately before the function is called (schema in Appendix~B.1, builder prompt in
Appendix~B.3, example in the left plate of Appendix Figure~B1).

\subsection{Contrastive Skill Memory}
\label{sec:skill_memory}

Skill Cards encode transferable procedural rules above individual function
syntax. For each reference task, \methodname{} compares the source trajectories
through the contrast packet---not to summarize every run, but to locate the
decision boundary that explains the outcome contrast. For example, when
failing agents guess an underspecified variant and execute a write while a
successful agent completes the safe branch and asks, the resulting card blocks
the ambiguous write until clarification. Supplementary Figure~B2 traces this
rule from cross-model disagreement to successful reuse on a new task; rendered
card formats appear in Supplementary Figure~B1.

\paragraph{Reflector.}
Given a contrast packet, the relevant current cards, and an
ability-specific focus, the Reflector isolates the decision boundary where
same-task trajectories diverged, attributes the success, failure, recovery,
or efficiency contrast, and abstracts it one level above the scenario into a
reusable rule with an explicit completion condition. Two constraints keep the output grounded: every field
must be supported by packet evidence, and if all source trajectories fail, the
Reflector may emit only diagnostic or recovery deltas rather than inventing a
success recipe (full protocol and focus blocks in Appendix~B.4).

The Reflector returns at most three candidate deltas per the Skill Delta
schema in Appendix~B.1; the right plate of Appendix Figure~B1 shows a
rendered card.

\paragraph{Curator.}
Candidate deltas are not written directly into the bank: for each delta, the
Curator (Appendix~B.5) retrieves the most relevant existing cards from the
same ability bank and decides the smallest valid edit among \texttt{ADD},
\texttt{PATCH}, \texttt{MERGE}, \texttt{NARROW}, and \texttt{NOOP} (patch
schema in Appendix~B.1). The Curator preserves semantic closure: it may combine, compress, narrow, or
rephrase information supported by the delta, the targeted cards, or related
Function Cards, but can introduce no new rule, failure mode, example, function, or
completion condition---preventing drift toward generic advice and
narrow-evidence rewrites.

The resulting Skill Card keeps the delta's compact runtime fields; provenance,
bookkeeping artifacts, and curator reasoning remain outside runtime memory. The five final ability banks contain 256 Skill Cards; across
them the Curator makes 394 localized decisions: 262 additions and 132
patches, merges, trigger narrowings, or no-ops (Appendix Figure~A1). Thus bank
growth is governed by evidence-sensitive local edits rather than append-only
accumulation.

\subsection{Retrieval and Runtime Injection}
\label{sec:retrieval_injection}

At evaluation time, the task's ability label selects which skill bank is eligible; when no label is available (as in open-ended deployment), retrieval simply retrieve over all banks. Within the selected bank, the reported GAIA2/ARE retriever ranks active Skill Cards by BM25 relevance with a capped soft app-domain overlap bonus and a domain-mismatch penalty.
These domain terms affect ranking rather than hard-gating cross-family transfer, so app-independent procedures, such as requesting clarification before an ambiguous write, remain
retrievable across application families. Appendix~A.8 gives the frozen benchmark-specific scores and selection caps. Function Cards follow a separate path: the top globally relevant cards are injected at task start; when the runtime identifies a proposed call, the matching card surfaces immediately before it as a just-in-time reminder of contract, returns, side effects, and common mistakes.

Injected memory is soft procedural guidance: current app observations remain
ground truth, and the target must not copy stored entities, dates,
identifiers, or final answers (complete task-start and pre-tool templates in
Appendix~B.6).

\paragraph{Validation.}\label{sec:validation}
\methodname{} validates memory at three levels: schema validation rejects any
card or patch unsupported by the delta, the targeted cards, or related
Function Cards; privacy validation rejects contact details, long identifiers,
raw judge prose, and copied scenario-specific values; and retrieval preview
verifies that held-out queries retrieve plausible cards from the correct bank
without same-scenario leakage. Without these checks, a bank can look strong by memorizing tasks or
replaying values.

\section{Experiments}
\label{sec:experiments}

\subsection{Experimental Setup}

We evaluate \methodname{} on \gaia{}/\are{}, the \gaia{} benchmark
executed in the Agents Research Environments (\are{})
runtime~\citep{froger2026gaia2,are}, across five abilities (Execution,
Search, Ambiguity, Adaptability, Time), with 40 reference and 40 held-out
tasks per ability. The bank is built from three heterogeneous source models (GPT-5.5, Claude
Sonnet 4.6, and DeepSeek V4 Pro) and evaluated on these target families plus
Qwen3.7 Plus, reserved as the unseen-transfer target with no contributed
trajectories. Each source target also gets a self-memory control: the same pipeline on
only its own reference trajectories. All agents are frontier-level, so gains are headroom on top of
already-capable agents. Construction consumes 600 offline
source trajectories ($5$ abilities $\times$ $40$ tasks $\times$ $3$ models)
and yields a frozen bank of 256 Skill Cards and 74 global Function Cards
(Appendix Figure~A1).
All experiments use the default runtime except the Time ability, where
wrapper-timestamp confounds require the normalized runtime for all
conditions.
\methodname{} retrieves up to three Skill Cards at start of the task and supplies
Function Card guidance before identifiable tool calls; every main evaluation compares the three conditions on the same split
(frozen configuration: Appendix Table~A1). Our primary metric is task success rate (the fraction of tasks passing the
environment verifier); efficiency is measured in \emph{agent events}, the
number of agent-level trajectory events per scenario.

To compare against published agent-memory baselines, we evaluate offline
Agent Workflow Memory (AWM)~\citep{wang2024awm}, ACE~\citep{zhang2025ace},
and raw trajectory retrieval on GPT-5.5 over Execution, Search, and
Ambiguity, testing whether the gains reduce to trajectory replay, induced
workflows, or an evolving playbook. All conditions share the three-model source pool and held-out scenarios
(adaptation and accounting details in Appendices~A.6--A.7).

\subsection{Main Held-Out Results}

\begin{table*}[t]
\centering
\setlength{\tabcolsep}{4pt}
{\small
\begin{tabular}{llrrrrrr}
\toprule
\textbf{Target} & \textbf{Method} &
\textbf{Exec.} & \textbf{Search} & \textbf{Ambig.} &
\textbf{Adapt.} & \textbf{Time} & \textbf{Overall} \\
\midrule
\multicolumn{8}{l}{\textit{Source target models}} \\
\addlinespace[1pt]
\multirow{3}{*}{GPT-5.5}
& No memory & 47.5 & 52.5 & 12.5 & 17.5 & 7.5 & 27.5 \\
& Self memory & 72.5 & 92.5 & 40.0 & 35.0 & 12.5 & 50.5 (+23.0) \\
& \textbf{\methodname{}} & \textbf{80.0} & \textbf{100.0} & \textbf{52.5}
& \textbf{55.0} & \textbf{17.5} & \textbf{61.0} (+33.5) \\
\midrule
\multirow{3}{*}{Claude Sonnet}
& No memory & 40.0 & 57.5 & 17.5 & 22.5 & 2.5 & 28.0 \\
& Self memory & 60.0 & 80.0 & 52.5 & \textbf{52.5} & \textbf{10.0} & 51.0 (+23.0) \\
& \textbf{\methodname{}} & \textbf{75.0} & \textbf{85.0} & \textbf{57.5}
& 42.5 & 2.5 & \textbf{52.5} (+24.5) \\
\midrule
\multirow{3}{*}{DeepSeek V4 Pro}
& No memory & 35.0 & 47.5 & 12.5 & 17.5 & 2.5 & 23.0 \\
& Self memory & 52.5 & \textbf{77.5} & 32.5 & 35.0 & 2.5 & 40.0 (+17.0) \\
& \textbf{\methodname{}} & \textbf{75.0} & \textbf{77.5} & \textbf{55.0}
& \textbf{52.5} & 2.5 & \textbf{52.5} (+29.5) \\
\midrule
\multirow{3}{*}{Source macro}
& No memory & 40.8 & 52.5 & 14.2 & 19.2 & 4.2 & 26.2 \\
& Self memory & 61.7 & 83.3 & 41.7 & 40.8 & \textbf{8.3} & 47.2 (+21.0) \\
& \textbf{\methodname{}} & \textbf{76.7} & \textbf{87.5} & \textbf{55.0}
& \textbf{50.0} & 7.5 & \textbf{55.3} (+29.2) \\
\midrule
\multicolumn{8}{l}{\textit{Unseen target model}} \\
\addlinespace[1pt]
\multirow{2}{*}{Qwen3.7 Plus}
& No memory & 25.0 & 47.5 & 5.0 & 15.0 & 0.0 & 18.5 \\
& \textbf{\methodname{}} & \textbf{50.0} & \textbf{67.5} & \textbf{20.0}
& \textbf{40.0} & 0.0 & \textbf{35.5} (+17.0) \\
\bottomrule
\end{tabular}
}
\caption{
Held-out success rates (\%) on GAIA2/ARE (40 tasks per ability and target).
Self memory applies the same pipeline to only the target's own reference
trajectories; \methodname{} uses the shared three-model bank. Qwen3.7 Plus is unseen during construction,
hence no self-memory row. Bold: best per column; parentheses: absolute gain
over no memory.
}
\label{tab:main_heldout}
\end{table*}

Table~\ref{tab:main_heldout} shows that \methodname{} more than doubles
held-out success on tasks unseen during construction, raising the source-target
macro from 26.2\% to 55.3\% with consistent per-target gains
(GPT-5.5 +33.5, Claude Sonnet +24.5, DeepSeek V4 Pro +29.5). Self memory, which runs the same pipeline on one model's trajectories, is already
effective (47.2\% source-target macro versus 26.2\% without), and the
shared bank adds 8.2 points on top, reaching 55.3\% and transferring to the
unseen Qwen3.7 Plus ($18.5\!\to\!35.5$):
\methodname{} distills reusable guidance, not model-specific behavior. All four non-temporal abilities improve (Execution +33.1, Search +31.2,
Ambiguity +34.4, Adaptability +29.4) with Analysis tracing the gains to the
memory contents.

Time ability remains hardest: obligations must survive asynchronous notifications
and land within narrow verifier windows against a simulated clock. Memory
still helps behaviorally (GPT-5.5 rises from $3/40$ to $7/40$; hung runs drop sharply,
Appendix~A.2), but self and contrastive
memory remain close across the three source targets ($10/120$ versus
$9/120$): cross-model evidence alone does not resolve temporal control. This marks the limit of text-injected procedural memory: it reliably
prevents the failures it has encoded; errors beyond the source evidence call
for a runtime controller, not richer recall.

Across all paired held-out runs, the gains are statistically significant (232
failure-to-pass versus 23 pass-to-fail flips; McNemar $\chi^2=171.3$,
$p \ll 0.001$) \citep{mcnemar1947sampling}. Across all five abilities,
\methodname{} reduces the macro-average trajectory length from 41.1 to 35.5
agent events per task ($-$5.6 events; $-$13.6\%).

\subsection{Cross-Benchmark Validation on AppWorld}
\label{sec:appworld}

\begin{table*}[t]
\centering
\setlength{\tabcolsep}{5pt}
{\small
\begin{tabular}{llrrrrrr}
\toprule
& & \multicolumn{3}{c}{\textbf{Test-Normal}} & \multicolumn{3}{c}{\textbf{Test-Challenge}} \\
\cmidrule(lr){3-5}\cmidrule(lr){6-8}
\textbf{Target} & \textbf{Method} & \textbf{TGC} & \textbf{SGC} & \textbf{$\Delta$ Base}
& \textbf{TGC} & \textbf{SGC} & \textbf{$\Delta$ Base} \\
\midrule
\multirow{3}{*}{DeepSeek-V3.1}
& No memory & 51.2 & 30.4 & -- & 56.8 & 38.1 & -- \\
& Self memory & 56.0 & 35.7 & +4.8 & 64.3 & 46.8 & +7.4 \\
& \textbf{\methodname{}} & \textbf{76.2} & \textbf{58.9} & \textbf{+25.0}
& \textbf{70.0} & \textbf{56.1} & \textbf{+13.2} \\
\midrule
\multirow{3}{*}{Qwen3.6-Flash}
& No memory & 63.7 & 42.9 & -- & 46.0 & 31.7 & -- \\
& Self memory & 73.2 & 58.9 & +9.5 & 50.8 & 37.4 & +4.8 \\
& \textbf{\methodname{}} & \textbf{80.4} & \textbf{71.4} & \textbf{+16.7}
& \textbf{59.5} & \textbf{47.5} & \textbf{+13.4} \\
\midrule
\multirow{3}{*}{GPT-4.1-mini}
& No memory & 48.2 & 25.0 & -- & 25.2 & 7.2 & -- \\
& Self memory & 60.7 & 39.3 & +12.5 & 36.2 & 18.0 & +11.0 \\
& \textbf{\methodname{}} & \textbf{68.5} & \textbf{50.0} & \textbf{+20.2}
& \textbf{42.7} & \textbf{23.0} & \textbf{+17.5} \\
\midrule
\multirow{3}{*}{Target macro}
& No memory & 54.4 & 32.7 & -- & 42.7 & 25.7 & -- \\
& Self memory & 63.3 & 44.6 & +8.9 & 50.4 & 34.1 & +7.8 \\
& \textbf{\methodname{}} & \textbf{75.0} & \textbf{60.1} & \textbf{+20.6}
& \textbf{57.4} & \textbf{42.2} & \textbf{+14.7} \\
\bottomrule
\end{tabular}
}
\caption{
AppWorld task (TGC) and scenario (SGC) goal completion (\%) on both public
test splits. Self memory applies the pipeline to each target's own train-split
trajectories (self-curated); \methodname{} uses the shared three-model bank;
all memories are frozen. Bold: best. $\Delta$ Base: TGC gain over no memory.
}
\label{tab:appworld}
\end{table*}

We further validate the same recipe on AppWorld~\citep{trivedi2024appworld},
which changes nearly everything about the execution regime: the agent writes
Python in a stateful REPL---orchestrating roughly 450 APIs across nine
applications, consulting live documentation, and finalizing through an explicit
completion call. State-based unit tests check required effects and collateral damage;
results are task (TGC) and scenario (SGC) goal completion on both public
splits (Test-Normal: 168 tasks / 56 scenarios; Test-Challenge: 417 / 139).

Because frontier models leave little headroom on much of AppWorld, we
deliberately evaluate three mid-tier agents, including DeepSeek-V3.1, Qwen3.6-Flash,
and GPT-4.1-mini, so measurement again reflects procedural headroom, not raw
capability. The same three agents serve as sources, each attempting every
training-split task once; the same-task triples are contrasted exactly as on
GAIA2. Two construction elements are benchmark-aware rather than benchmark-tuned. 
Because AppWorld agents can read live documentation at any point,
Function/API Cards obey a \emph{doc-delta} rule: they record only
evidence-backed behavior beyond the documentation, such as exception
signatures, argument constraints, and pagination, and never restate it.
In addition, a deterministic task signature (answer extraction,
bulk/multi-write, state write, cross-app, default) selects the reflection
focus, replacing GAIA2's ability focus. Construction uses GPT-4.1-mini,
the weakest of the trio, as \reflector{} and \curator{}, so gains cannot
come from a stronger construction model;
each self-memory control is fully self-contained, the target curating its own
trajectories. All banks are frozen; retrieved cards (at most three Skill and two
Function/API Cards, about 1.1k tokens) are injected once at task start as soft guidance under an unchanged
prompt. Construction
yields 32 Skill and 24 Function/API Cards over 60 Curator decisions (Appendix
Figure~A1; prompt suite and injection contract in Appendices~B.7--B.11).

Table~\ref{tab:appworld} shows a consistent ordering across every target,
split, and metric: \methodname{} outperforms both no memory and
target-specific self memory. Averaged over the three targets, Test-Normal TGC
rises from $54.4\%$ without memory to $63.3\%$ with self memory and $75.0\%$
with \methodname{}; on Test-Challenge the progression is
$42.7\%\!\to\!50.4\%\!\to\!57.4\%$, with per-target gains of $+16.7$--$+25.0$ (Test-Normal) and
$+13.2$--$+17.5$ (Test-Challenge). The pattern is strongest for SGC, which requires every task
in a scenario to pass: the cross-model bank reaches $60.1\%$ and $42.2\%$
versus $44.6\%$ and $34.1\%$ for self memory---informative, because
scenario completion rewards exactly the discipline (complete enumeration
before writing, verification before finalization) the cards encode. The consistent margin over self memory shows that heterogeneous
trajectories add value beyond retention, without ensembling or parameter
updates.

\section{Ablations and Analysis}
\label{sec:ablations}

We run cost-controlled ablations isolating source diversity, curation, and
memory type on the three most diagnostic abilities (Search, Ambiguity, and
Execution) on the held-out split with the frozen retrieval path. Table~\ref{tab:components} reports source and component ablations;
Figure~\ref{fig:memory_baselines} compares prior memory systems.

\begin{table}[t]
\centering
\setlength{\tabcolsep}{4pt}
{\small
\begin{tabularx}{\linewidth}{@{}>{\raggedright\arraybackslash}Xrrrr@{}}
\toprule
\textbf{Memory variant} & \textbf{Exec.} & \textbf{Search} & \textbf{Ambig.} & \textbf{Macro} \\
\midrule
No memory & 47.5 & 52.5 & 12.5 & 37.5 \\
\midrule
GPT-5.5 self memory & 72.5 & 92.5 & 40.0 & 68.3 \\
GPT-5.5 multi-rollout & 75.0 & 92.5 & 42.5 & 70.0 \\
\midrule
Function Cards only & 67.5 & 85.0 & 30.0 & 60.8 \\
Skill Cards only & 60.0 & 100.0 & 40.0 & 66.7 \\
\midrule
\methodname{} (three models) & \textbf{80.0} & \textbf{100.0} & \textbf{52.5} & \textbf{77.5} \\
\bottomrule
\end{tabularx}
}
\caption{GPT-5.5 held-out source and component ablations. Multi-rollout uses
three GPT-5.5 trajectories, matching \methodname{}'s count; single-card rows
use the shared bank and frozen retrieval path. Macro averages the three
abilities.}
\vspace{-10pt}
\label{tab:components}
\end{table}

\subsection{Source Diversity and Curation}

Self memory is not a third-party baseline but \methodname{} run
single-source, with the same normalization, typed cards, \curator{}, and
retrieval; its 26.2\%$\to$47.2\% source-target gain
(Table~\ref{tab:main_heldout}) measures the architecture alone; the shared
bank's 55.3\% (margins +10.5 on GPT-5.5, +12.5 on DeepSeek) isolates the
added value of heterogeneous contrast. To separate diversity from trajectory count, Table~\ref{tab:components}
adds a matched multi-rollout control with three GPT-5.5 trajectories per
task; it trails \methodname{} on every ability and on macro average (70.0
vs.\ 77.5): additional sampling alone does not explain the gain.

We isolate the Curator against an append-only variant committing every
Reflector delta as \textsc{Add}, with identical deltas and retrieval. On the four non-temporal abilities covered by this ablation, local curation removes 24.9\% of Skill Cards ($273\!\to\!205$) while
improving GPT-5.5 held-out macro success from 65.0\% to 71.9\% (Appendix Table~A4)---a smaller
bank, no ability regressing, strongest on Execution (+15.0), where redundant
write-coverage cards compete for fixed injection slots.

\subsection{Memory Types and Prior Memory Systems}

\begin{figure}[t]
\centering
\includegraphics[width=0.98\linewidth]{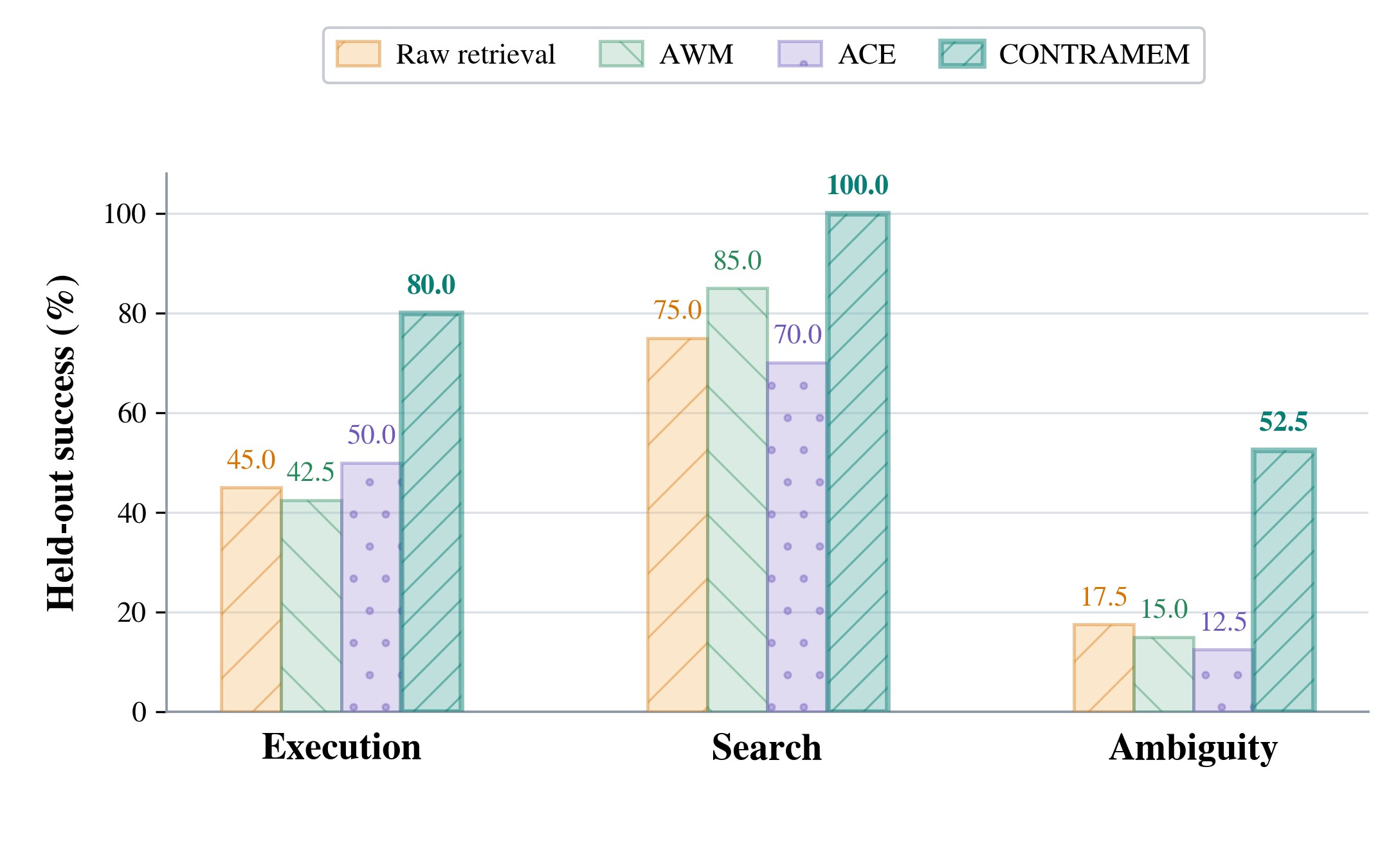}
\vspace{-10pt}
    \caption{\methodname{} versus trajectory- and playbook-based memory on
    GPT-5.5 held-out tasks. All methods share the same source pool and
    scenarios. Exact values: Appendix Table~A5.}
\label{fig:memory_baselines}
\end{figure}

The single-memory-type rows of Table~\ref{tab:components} isolate the two
levels: Skill Cards carry much of Search and Ambiguity, Function Cards matter
most where correctness depends on concrete writes (Execution), and their
combination attains the best macro average. The levels are complementary.

Figure~\ref{fig:memory_baselines} compares \methodname{} with
raw retrieval, offline AWM~\citep{wang2024awm}, and
ACE~\citep{zhang2025ace} under their native learning rules. AWM induces a
fixed workflow library from verifier-passing trajectories, whereas ACE turns
all 120 trajectories per ability into an evolving full playbook; neither uses
same-task contrasts, typed Function/Skill memory, or task-conditioned
retrieval (Appendices~A.6--A.7). 
AWM is the strongest prior baseline
(47.5\% macro; 17 vs.\ 5 discordant flips, $p{=}0.0169$), followed by raw
retrieval (45.8\%) and ACE (44.2\%). \methodname{} reaches 77.5\%: 30.0
points above AWM, with 38 favorable flips against 2 ($p{<}10^{-8}$),
showing that preserving decision boundaries is more useful than retaining
trace detail alone.
\paragraph{Analysis.}\label{sec:analysis}
The largest gains come where the no-memory agent has enough evidence but
writes, asks, retries, or stops at the wrong boundary: in Ambiguity, retrieved
skills convert ``guess and write'' into ``complete safe branches and ask
before side effects'' (Supplementary Figure~B2); in Search, they identify when
the answer is determined; in Execution and Adaptability, they preserve
obligations. The 23 pass-to-fail regressions arise mainly from over-broad
clarification cards or caution around already-determined writes, motivating
conservative top-$k$ retrieval. Transfer to Qwen3.7 Plus
($18.5\!\to\!35.5$, with no contributed trajectories) further indicates that
shared-scenario divergence exposes task- and tool-level invariants rather than
source-specific detours.

\section{Conclusion}

We introduced \method{}, which turns same-task disagreement into procedural
supervision and distills it into a compact, locally curated bank of Function
and Skill Cards. It more than doubles held-out success rate on \gaia{}/\are{},
outperforms single-source, workflow, and playbook memory, transfers unchanged
to an unseen target, and improves three target agents on both AppWorld test
splits.

Our work admittedly has limitations: \method{} remains bounded by the
procedural failures represented in its source trajectories, and text-injected
memory alone does not resolve time-sensitive failures requiring persistent
state tracking and runtime control. Combining curated memory with mechanisms
for obligation tracking and asynchronous-state monitoring is a promising next
step.

\begingroup\small
\bibliography{references}

\begin{thebibliography}{35}
\providecommand{\natexlab}[1]{#1}

\bibitem[{Andrews et~al.(2025)Andrews, Benhalloum, Bertran et~al.}]{are}
Andrews, P.; Benhalloum, A.; Bertran, G. M.-T.; et~al. 2025.
\newblock ARE: Scaling Up Agent Environments and Evaluations.
\newblock \emph{arXiv preprint arXiv:2509.17158}.

\bibitem[{Besta et~al.(2024)Besta, Blach, Kubicek, Gerstenberger, Podstawski,
  Gianinazzi, Gajda, Lehmann, Niewiadomski, Nyczyk et~al.}]{besta2023got}
Besta, M.; Blach, N.; Kubicek, A.; Gerstenberger, R.; Podstawski, M.;
  Gianinazzi, L.; Gajda, J.; Lehmann, T.; Niewiadomski, H.; Nyczyk, P.; et~al.
  2024.
\newblock Graph of thoughts: Solving elaborate problems with large language
  models.
\newblock In \emph{Proceedings of the AAAI conference on artificial
  intelligence}, volume~38, 17682--17690.

\bibitem[{Cao et~al.(2026)Cao, Deng, Yu, Zhou, Liu, Ding, and
  Zhao}]{cao2026reme}
Cao, Z.; Deng, J.; Yu, L.; Zhou, W.; Liu, Z.; Ding, B.; and Zhao, H. 2026.
\newblock Remember me, refine me: A dynamic procedural memory framework for
  experience-driven agent evolution.
\newblock In \emph{Findings of the Association for Computational Linguistics:
  ACL 2026}, 16803--16822.

\bibitem[{Chang et~al.(2026)Chang, Wu, Wu, and Lin}]{chang2026memcollab}
Chang, Y.; Wu, Y.; Wu, Q.; and Lin, L. 2026.
\newblock MemCollab: Cross-Agent Memory Collaboration via Contrastive
  Trajectory Distillation.
\newblock \emph{arXiv preprint arXiv:2603.23234}.

\bibitem[{Chen et~al.(2020)Chen, Kornblith, Norouzi, and
  Hinton}]{chen2020simclr}
Chen, T.; Kornblith, S.; Norouzi, M.; and Hinton, G. 2020.
\newblock A Simple Framework for Contrastive Learning of Visual
  Representations.
\newblock In \emph{Proceedings of the 37th International Conference on Machine
  Learning}, volume 119 of \emph{Proceedings of Machine Learning Research},
  1597--1607. PMLR.

\bibitem[{{DeepSeek-AI} et~al.(2025){DeepSeek-AI}, Guo, Yang, Zhang, Song,
  Zhang, Xu, Zhu, Ma, Wang et~al.}]{guo2025deepseekr1}
{DeepSeek-AI}; Guo, D.; Yang, D.; Zhang, H.; Song, J.; Zhang, R.; Xu, R.; Zhu,
  Q.; Ma, S.; Wang, P.; et~al. 2025.
\newblock DeepSeek-R1: Incentivizing Reasoning Capability in LLMs via
  Reinforcement Learning.
\newblock \emph{arXiv preprint arXiv:2501.12948}.

\bibitem[{Fang et~al.(2026)Fang, Isahagian, Jayaram, Kumar, Muthusamy, Oum, and
  Thomas}]{fang2026trajectorymemory}
Fang, G.; Isahagian, V.; Jayaram, K.~R.; Kumar, R.; Muthusamy, V.; Oum, P.; and
  Thomas, G. 2026.
\newblock Trajectory-Informed Memory Generation for Self-Improving Agent
  Systems.
\newblock \emph{arXiv preprint arXiv:2603.10600}.

\bibitem[{Froger et~al.(2026)Froger, Andrews, Bettini, Budhiraja, Cabral, Do,
  Garreau, Gaya, Lauren{\c{c}}on, Lecanu, Malkan, Mekala, M{\'e}nard, Bertran,
  Piterbarg, Plekhanov, Rita, Rusakov, Vorotilov, Wang, Yu, Benhalloum, Mialon,
  and Scialom}]{froger2026gaia2}
Froger, R.; Andrews, P.; Bettini, M.; Budhiraja, A.; Cabral, R.~S.; Do, V.;
  Garreau, E.; Gaya, J.-B.; Lauren{\c{c}}on, H.; Lecanu, M.; Malkan, K.;
  Mekala, D.; M{\'e}nard, P.; Bertran, G. M.-T.; Piterbarg, U.; Plekhanov, M.;
  Rita, M.; Rusakov, A.; Vorotilov, V.; Wang, M.; Yu, I.; Benhalloum, A.;
  Mialon, G.; and Scialom, T. 2026.
\newblock Gaia2: Benchmarking LLM Agents on Dynamic and Asynchronous
  Environments.
\newblock \emph{arXiv preprint arXiv:2602.11964}.

\bibitem[{Guo et~al.(2026)Guo, Lin, Wang, Han, Hu, Ni, Wang, and
  Chen}]{lin2025seagent}
Guo, Y.; Lin, J.; Wang, H.; Han, Y.; Hu, S.; Ni, Z.; Wang, L.; and Chen, M.
  2026.
\newblock SE-agent: Self-evolution trajectory optimization in multi-step
  reasoning with LLM-based agents.
\newblock \emph{Advances in Neural Information Processing Systems}, 38:
  116314--116341.

\bibitem[{He et~al.(2019)He, Fan, Wu, Xie, and Girshick}]{he2020moco}
He, K.; Fan, H.; Wu, Y.; Xie, S.; and Girshick, R. 2019.
\newblock Momentum contrast for unsupervised visual representation learning.
\newblock \emph{arXiv preprint arXiv:1911.05722}.

\bibitem[{Li et~al.(2023)Li, Zhao, Yu, Song, Li, Yu, Li, Huang, and
  Li}]{li2023api}
Li, M.; Zhao, Y.; Yu, B.; Song, F.; Li, H.; Yu, H.; Li, Z.; Huang, F.; and Li,
  Y. 2023.
\newblock Api-bank: A comprehensive benchmark for tool-augmented llms.
\newblock In \emph{Proceedings of the 2023 conference on empirical methods in
  natural language processing}, 3102--3116.

\bibitem[{Liu et~al.(2025)Liu, Si, Narasimhan, and Yao}]{liu2025cer}
Liu, Y.; Si, C.; Narasimhan, K.~R.; and Yao, S. 2025.
\newblock Contextual experience replay for self-improvement of language agents.
\newblock In \emph{Proceedings of the 63rd Annual Meeting of the Association
  for Computational Linguistics (Volume 1: Long Papers)}, 14179--14198.

\bibitem[{Lu et~al.(2026)Lu, Deng, Zhang, Hu, Zhao, Tian, Mou, Lin, and
  Li}]{lu2026muon}
Lu, B.; Deng, Z.; Zhang, R.; Hu, B.; Zhao, Y.; Tian, Y.; Mou, C.; Lin, G.; and
  Li, X. 2026.
\newblock Muon-OGD: Muon-based spectral orthogonal gradient projection for LLM
  continual learning.
\newblock \emph{arXiv preprint arXiv:2605.08949}.

\bibitem[{Lu et~al.(2025)Lu, Holleis, Zhang, Aumayer, Nan, Bai, Ma, Ma, Li, Yin
  et~al.}]{lu2024toolsandbox}
Lu, J.; Holleis, T.; Zhang, Y.; Aumayer, B.; Nan, F.; Bai, H.; Ma, S.; Ma, S.;
  Li, M.; Yin, G.; et~al. 2025.
\newblock Toolsandbox: A stateful, conversational, interactive evaluation
  benchmark for llm tool use capabilities.
\newblock In \emph{Findings of the Association for Computational Linguistics:
  NAACL 2025}, 1160--1183.

\bibitem[{McNemar(1947)}]{mcnemar1947sampling}
McNemar, Q. 1947.
\newblock Note on the Sampling Error of the Difference Between Correlated
  Proportions or Percentages.
\newblock \emph{Psychometrika}, 12(2): 153--157.

\bibitem[{Ouyang et~al.(2026)Ouyang, Yan, Hsu, Chen, Jiang, Wang, Han, Le,
  Daruki, Tang, Tirumalashetty, Lee, Rofouei, Lin, Han, Lee, and
  Pfister}]{ouyang2026reasoningbank}
Ouyang, S.; Yan, J.; Hsu, I.-H.; Chen, Y.; Jiang, K.; Wang, Z.; Han, R.; Le,
  L.~T.; Daruki, S.; Tang, X.; Tirumalashetty, V.; Lee, G.; Rofouei, M.; Lin,
  H.; Han, J.; Lee, C.-Y.; and Pfister, T. 2026.
\newblock ReasoningBank: Scaling Agent Self-Evolving with Reasoning Memory.
\newblock \emph{arXiv preprint arXiv:2509.25140}.

\bibitem[{Packer et~al.(2023)Packer, Wooders, Lin, Fang, Patil, Stoica, and
  Gonzalez}]{packer2023memgpt}
Packer, C.; Wooders, S.; Lin, K.; Fang, V.; Patil, S.~G.; Stoica, I.; and
  Gonzalez, J.~E. 2023.
\newblock MemGPT: Towards LLMs as Operating Systems.
\newblock \emph{arXiv preprint arXiv:2310.08560}.

\bibitem[{Park et~al.(2023)Park, O'Brien, Cai, Morris, Liang, and
  Bernstein}]{park2023generativeagents}
Park, J.~S.; O'Brien, J.; Cai, C.~J.; Morris, M.~R.; Liang, P.; and Bernstein,
  M.~S. 2023.
\newblock Generative agents: Interactive simulacra of human behavior.
\newblock In \emph{Proceedings of the 36th annual acm symposium on user
  interface software and technology}, 1--22.

\bibitem[{Patil et~al.(2024)Patil, Zhang, Wang, and
  Gonzalez}]{patil2024gorilla}
Patil, S.~G.; Zhang, T.; Wang, X.; and Gonzalez, J.~E. 2024.
\newblock Gorilla: Large language model connected with massive apis.
\newblock \emph{Advances in Neural Information Processing Systems}, 37:
  126544--126565.

\bibitem[{Schick et~al.(2023)Schick, Dwivedi-Yu, Dess{\`\i}, Raileanu, Lomeli,
  Hambro, Zettlemoyer, Cancedda, and Scialom}]{schick2023toolformer}
Schick, T.; Dwivedi-Yu, J.; Dess{\`\i}, R.; Raileanu, R.; Lomeli, M.; Hambro,
  E.; Zettlemoyer, L.; Cancedda, N.; and Scialom, T. 2023.
\newblock Toolformer: Language models can teach themselves to use tools.
\newblock \emph{Advances in neural information processing systems}, 36:
  68539--68551.

\bibitem[{Shinn et~al.(2023)Shinn, Cassano, Gopinath, Narasimhan, and
  Yao}]{shinn2023reflexion}
Shinn, N.; Cassano, F.; Gopinath, A.; Narasimhan, K.; and Yao, S. 2023.
\newblock Reflexion: Language agents with verbal reinforcement learning.
\newblock \emph{Advances in neural information processing systems}, 36:
  8634--8652.

\bibitem[{Tang et~al.(2025)Tang, Qin, Peng, Zhou, Shao, Du, Wei, Xia, Wu, Zhu,
  Zhang, Liu, Wang, Hong, Wu, Cheng, Wang, and Zhou}]{tang2025agentkb}
Tang, X.; Qin, T.; Peng, T.; Zhou, Z.; Shao, D.; Du, T.; Wei, X.; Xia, P.; Wu,
  F.; Zhu, H.; Zhang, G.; Liu, J.; Wang, X.; Hong, S.; Wu, C.; Cheng, H.; Wang,
  C.; and Zhou, W. 2025.
\newblock Agent KB: Leveraging Cross-Domain Experience for Agentic Problem
  Solving.
\newblock \emph{arXiv preprint arXiv:2507.06229}.

\bibitem[{Trivedi et~al.(2024)Trivedi, Khot, Hartmann, Manku, Dong, Li, Gupta,
  Sabharwal, and Balasubramanian}]{trivedi2024appworld}
Trivedi, H.; Khot, T.; Hartmann, M.; Manku, R.; Dong, V.; Li, E.; Gupta, S.;
  Sabharwal, A.; and Balasubramanian, N. 2024.
\newblock Appworld: A controllable world of apps and people for benchmarking
  interactive coding agents.
\newblock In \emph{Proceedings of the 62nd Annual Meeting of the Association
  for Computational Linguistics (Volume 1: Long Papers)}, 16022--16076.

\bibitem[{van~den Oord, Li, and Vinyals(2018)}]{oord2018cpc}
van~den Oord, A.; Li, Y.; and Vinyals, O. 2018.
\newblock Representation Learning with Contrastive Predictive Coding.
\newblock \emph{arXiv preprint arXiv:1807.03748}.

\bibitem[{Wang et~al.(2023{\natexlab{a}})Wang, Xie, Jiang, Mandlekar, Xiao,
  Zhu, Fan, and Anandkumar}]{wang2023voyager}
Wang, G.; Xie, Y.; Jiang, Y.; Mandlekar, A.; Xiao, C.; Zhu, Y.; Fan, L.; and
  Anandkumar, A. 2023{\natexlab{a}}.
\newblock Voyager: An Open-Ended Embodied Agent with Large Language Models.
\newblock \emph{arXiv preprint arXiv:2305.16291}.

\bibitem[{Wang et~al.(2023{\natexlab{b}})Wang, Dong, Cheng, Liu, Yan, Gao, and
  Wei}]{wang2023augmenting}
Wang, W.; Dong, L.; Cheng, H.; Liu, X.; Yan, X.; Gao, J.; and Wei, F.
  2023{\natexlab{b}}.
\newblock Augmenting language models with long-term memory.
\newblock \emph{Advances in Neural Information Processing Systems}, 36:
  74530--74543.

\bibitem[{Wang et~al.(2024)Wang, Mao, Fried, and Neubig}]{wang2024awm}
Wang, Z.~Z.; Mao, J.; Fried, D.; and Neubig, G. 2024.
\newblock Agent Workflow Memory.
\newblock \emph{arXiv preprint arXiv:2409.07429}.

\bibitem[{Xiong et~al.(2025)Xiong, Lin, Xie, He, Tang, Lakkaraju, and
  Xiang}]{xiong2025memorymanagement}
Xiong, Z.; Lin, Y.; Xie, W.; He, P.; Tang, J.; Lakkaraju, H.; and Xiang, Z.
  2025.
\newblock How Memory Management Impacts LLM Agents: An Empirical Study of
  Experience-Following Behavior.
\newblock \emph{arXiv preprint arXiv:2505.16067}.

\bibitem[{Xu et~al.(2026)Xu, Liang, Mei, Gao, Tan, and Zhang}]{xu2025mem}
Xu, W.; Liang, Z.; Mei, K.; Gao, H.; Tan, J.; and Zhang, Y. 2026.
\newblock A-mem: Agentic memory for llm agents.
\newblock \emph{Advances in Neural Information Processing Systems}, 38:
  17577--17604.

\bibitem[{Yang et~al.(2024)Yang, Yu, Zhang, Cao, Xu, Zhang, Gonzalez, and
  Cui}]{yang2024buffer}
Yang, L.; Yu, Z.; Zhang, T.; Cao, S.; Xu, M.; Zhang, W.; Gonzalez, J.~E.; and
  Cui, B. 2024.
\newblock Buffer of thoughts: Thought-augmented reasoning with large language
  models.
\newblock \emph{Advances in Neural Information Processing Systems}, 37:
  113519--113544.

\bibitem[{Yao et~al.(2023)Yao, Yu, Zhao, Shafran, Griffiths, Cao, and
  Narasimhan}]{yao2023bot}
Yao, S.; Yu, D.; Zhao, J.; Shafran, I.; Griffiths, T.; Cao, Y.; and Narasimhan,
  K. 2023.
\newblock Tree of thoughts: Deliberate problem solving with large language
  models.
\newblock \emph{Advances in neural information processing systems}, 36:
  11809--11822.

\bibitem[{Yao et~al.(2022)Yao, Zhao, Yu, Shafran, Narasimhan, and
  Cao}]{yao2022react}
Yao, S.; Zhao, J.; Yu, D.; Shafran, I.; Narasimhan, K.~R.; and Cao, Y. 2022.
\newblock React: Synergizing reasoning and acting in language models.
\newblock In \emph{NeurIPS 2022 Foundation Models for Decision Making
  Workshop}.

\bibitem[{Zhang et~al.(2026)Zhang, Lin, Wu, Sun, Li, Li, and
  Peng}]{zhang2026faultymemory}
Zhang, D.; Lin, Y.; Wu, Z.; Sun, Y.; Li, B.; Li, D.; and Peng, H. 2026.
\newblock Useful Memories Become Faulty When Continuously Updated by LLMs.
\newblock \emph{arXiv preprint arXiv:2605.12978}.

\bibitem[{Zhang et~al.(2025)Zhang, Hu, Upasani, Ma, Hong, Kamanuru, Rainton,
  Wu, Ji, Li, Thakker, Zou, and Olukotun}]{zhang2025ace}
Zhang, Q.; Hu, C.; Upasani, S.; Ma, B.; Hong, F.; Kamanuru, V.; Rainton, J.;
  Wu, C.; Ji, M.; Li, H.; Thakker, U.; Zou, J.; and Olukotun, K. 2025.
\newblock Agentic Context Engineering: Evolving Contexts for Self-Improving
  Language Models.
\newblock \emph{arXiv preprint arXiv:2510.04618}.

\bibitem[{Zhao et~al.(2024)Zhao, Huang, Xu, Lin, Liu, and
  Huang}]{zhao2023expel}
Zhao, A.; Huang, D.; Xu, Q.; Lin, M.; Liu, Y.-J.; and Huang, G. 2024.
\newblock Expel: Llm agents are experiential learners.
\newblock In \emph{Proceedings of the AAAI Conference on Artificial
  Intelligence}, volume~38, 19632--19642.

\end{thebibliography}
\endgroup

\clearpage
\noindent This appendix accompanies the paper \emph{CONTRAMEM: Learning Self-Evolving Procedural
Memory from Contrasting Multi-Model Trajectories}. Appendix~A provides additional experimental
detail and the formal pipeline; Appendix~B provides memory schemas, rendered
card examples, and the complete construction and runtime prompts for both
benchmarks. Table and figure numbers match the references used in the main
paper.

\appendix
\setcounter{table}{0}
\setcounter{figure}{0}
\renewcommand{\thetable}{A\arabic{table}}
\renewcommand{\thefigure}{A\arabic{figure}}

\section{Additional Experimental Detail}
\label{app:additional}

\subsection{A.1\quad Memory-Bank Evolution and Frozen Configuration}
\label{app:eval_config}

\begin{figure*}[t!]
  \centering
  \includegraphics[width=0.92\textwidth]{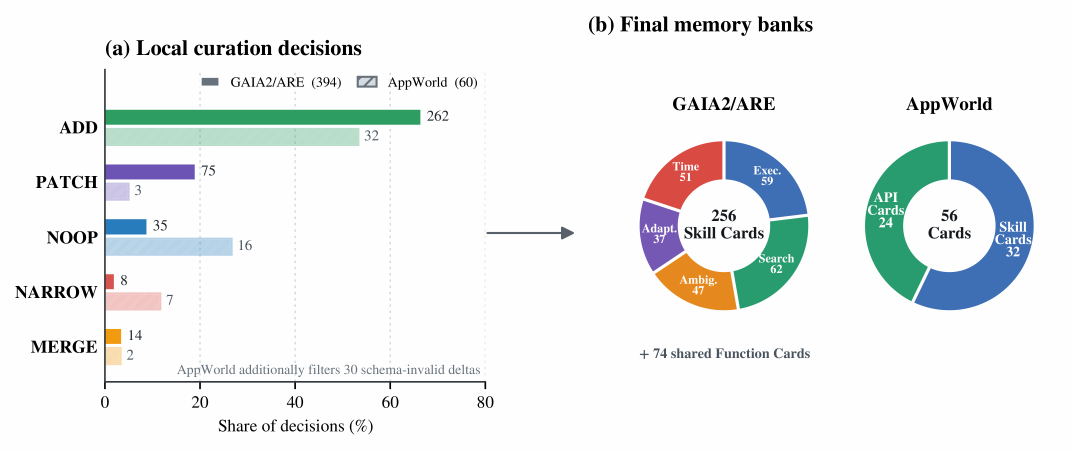}
  \captionof{figure}{Memory-bank evolution across benchmarks.
  (a) Distribution of localized Skill \curator{} decisions; numbers at bar
  ends are counts. Solid bars denote \gaia{}/\are{} (394 decisions) and
  hatched bars AppWorld (60). AppWorld additionally filters 30 schema-invalid
  candidate deltas before curation.
  (b) Final banks: \gaia{}/\are{} contains 256 ability-specific Skill Cards
  plus 74 shared Function Cards; AppWorld contains 32 Skill Cards and 24
  Function/API Cards.}
  \label{fig:self_evolution_operations}
  \vspace{0.3em}
  {\small
  \begin{tabularx}{\textwidth}{@{}lX@{}}
  \toprule
  \textbf{Item} & \textbf{Configuration} \\
  \midrule
  Dataset & GAIA2/ARE dynamic app-agent tasks. \\
  Split & Balanced seed 20260610; 40 reference and 40 held-out tasks per ability. \\
  Abilities & Execution, Search, Ambiguity, Adaptability, and Time. \\
  Memory source agents & GPT-5.5, Claude Sonnet 4.6, and DeepSeek V4 Pro. \\
  Target agents & GPT-5.5, Claude Sonnet 4.6, DeepSeek V4 Pro, and Qwen3.7 Plus. \\
  Memory construction model & GPT-5.5 with high reasoning effort for Function Card construction, Skill reflection, and Skill curation. \\
  Final memory bank & 256 Skill Cards (including the 51-card Time bank) and 74 global Function Cards. \\
  GPT-5.5 target setting & Default evaluation setting; no explicit extended-thinking override at runtime. \\
  Claude Sonnet target setting & High/hard extended-thinking setting for Claude Sonnet 4.6 target runs. \\
  Verifier & GAIA2 verifier with GPT-5.5 judge where LLM judgment is required. \\
  Runtime & Standard GAIA2/ARE integration for the four non-temporal abilities; Time uses the temporal runtime described below. \\
  Skill retrieval & BM25 over Skill Card title, trigger, tags, and core rule, with a soft app-family prior. \\
  Skill injection & Up to three task-start Skill Cards; Time may reserve up to two slots for ability-universal cards. \\
  Function injection & Retrieved Function Cards plus pre-tool function guidance; pre-tool guidance is disabled for Time. \\
  Memory policy & Retrieved memories are soft guidance; live environment observations remain ground truth. \\
  Timeout & 300s per-scenario runner timeout; 900s for Time under the revised temporal runtime (both conditions). \\
  Concurrency & Two parallel scenarios unless otherwise stated in a run manifest. \\
  \bottomrule
  \end{tabularx}
  }
  \captionof{table}{Frozen evaluation configuration used for the main results.}
  \label{tab:eval_config}
\end{figure*}

Figure~\ref{fig:self_evolution_operations} summarizes how the two memory banks
evolve under localized curation, and Table~\ref{tab:eval_config} records the
frozen configuration used for all main-table runs.

Time uses a benchmark-specific temporal adapter because asynchronous
obligations are evaluated against a simulated scenario clock rather than the
wrapper clock. The adapter exposes authoritative simulated time and emits
bounded, task-blind tick notifications so that pending obligations can resume.
For this ability, pre-tool guidance is disabled; up to two
ability-universal Skill Cards may be fixed at task start, and one finalization
card may be repeated near the final timer boundary. The tick mechanism itself
does not inspect task content or prescribe an application action. The adapter
also repairs a wrapper artifact in which a run would terminate on an empty
final boundary while obligations remain pending; this repair, like the tick
mechanism, is part of the shared Time image and applies identically to both
conditions. Task-start Function Cards for Time additionally pass a verb-intent
relevance filter, identical across the memory conditions.

\subsection{A.2\quad Trajectory Efficiency}
\label{app:efficiency}

Table~\ref{tab:heldout_efficiency} and Figure~\ref{fig:heldout_agent_events}
report per-ability agent-event statistics for the held-out evaluation
summarized in the main text. For Time, both conditions use the normalized
temporal runtime; neutral timer ticks make these traces longer in absolute
terms. Beyond the aggregate reduction, memory
also repairs stalled behavior on Time: hung runs fall from 1 to 0 for Claude
Sonnet and from 13 to 6 for DeepSeek V4 Pro, although many resumed executions
still fail at later steps absent from the source evidence.

\begin{table}[b]
\centering
{\small
\begin{tabular}{lrrr}
\toprule
\textbf{Ability} &
\textbf{Avg No} &
\textbf{Avg Ours} &
\textbf{$\Delta$} \\
\midrule
Exec          & 30.3 & 31.1 & +0.8 \\
Search        & 55.6 & 40.2 & \textbf{-15.5} \\
Ambig         & 29.5 & 24.3 & \textbf{-5.2} \\
Adapt         & 19.0 & 20.7 & +1.6 \\
Time          & 70.9 & 61.0 & \textbf{-9.9} \\
\midrule
\textbf{Macro} & \textbf{41.1} & \textbf{35.5} & \textbf{-5.6} \\
\bottomrule
\end{tabular}
}
\caption{
Agent-event efficiency on GAIA2/ARE held-out tasks, paired by scenario.
Negative $\Delta$ means that \methodname{} executes the task with fewer
agent-level events.
}
\label{tab:heldout_efficiency}
\end{table}

\subsection{A.3\quad Reference-Split Results}
\label{app:reference_results}

\begin{figure*}[t!]
  \centering
  \includegraphics[width=0.88\textwidth]{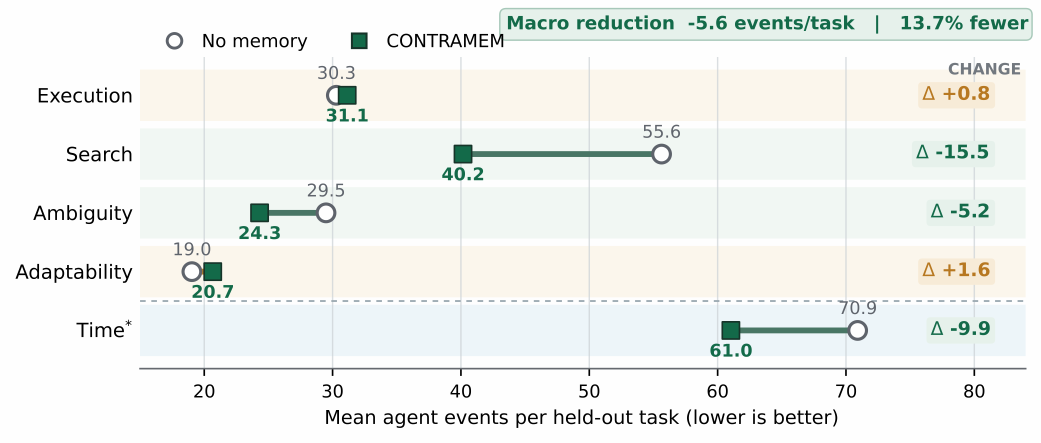}
  \captionof{figure}{Mean agent events per held-out task, paired by scenario. Lower is
  better. Small increases on Execution and Adaptability reflect additional
  necessary work completed by successful memory-conditioned agents.}
  \label{fig:heldout_agent_events}
  \vspace{0.2em}
  {\small
  \begin{tabular}{llrrrrrr}
  \toprule
  \textbf{Target} & \textbf{Method} &
  \textbf{Exec.} & \textbf{Search} & \textbf{Ambig.} &
  \textbf{Adapt.} & \textbf{Time} & \textbf{Overall} \\
  \midrule
  \multirow{2}{*}{GPT-5.5}
  & No memory & 45.0 & 52.5 & 10.0 & 17.5 & 2.5 & 25.5 \\
  & \methodname{} & \textbf{70.0} (+25.0)
  & \textbf{92.5} (+40.0)
  & \textbf{60.0} (+50.0)
  & \textbf{35.0} (+17.5)
  & \textbf{10.0} (+7.5)
  & \textbf{53.5} (+28.0) \\
  \midrule
  \multirow{2}{*}{Claude Sonnet}
  & No memory & 37.5 & 57.5 & 20.0 & 20.0 & 2.5 & 27.5 \\
  & \methodname{} & \textbf{70.0} (+32.5)
  & \textbf{70.0} (+12.5)
  & \textbf{52.5} (+32.5)
  & \textbf{50.0} (+30.0)
  & 0.0 (-2.5)
  & \textbf{48.5} (+21.0) \\
  \midrule
  \multirow{2}{*}{DeepSeek V4 Pro}
  & No memory & 32.5 & 47.5 & 12.5 & 17.5 & 0.0 & 22.0 \\
  & \methodname{} & \textbf{72.5} (+40.0)
  & \textbf{72.5} (+25.0)
  & \textbf{42.5} (+30.0)
  & \textbf{47.5} (+30.0)
  & 0.0 (+0.0)
  & \textbf{47.0} (+25.0) \\
  \midrule
  \multirow{2}{*}{Qwen3.7 Plus}
  & No memory & 12.5 & 42.5 & 7.5 & 17.5 & 0.0 & 16.0 \\
  & \methodname{} & \textbf{37.5} (+25.0)
  & \textbf{60.0} (+17.5)
  & \textbf{20.0} (+12.5)
  & \textbf{32.5} (+15.0)
  & 0.0 (+0.0)
  & \textbf{30.0} (+14.0) \\
  \midrule
  \multicolumn{2}{r}{\textbf{Aggregate}}
  & 31.9 $\rightarrow$ \textbf{62.5}
  & 50.0 $\rightarrow$ \textbf{73.8}
  & 12.5 $\rightarrow$ \textbf{43.8}
  & 18.1 $\rightarrow$ \textbf{41.3}
  & 1.3 $\rightarrow$ \textbf{2.5}
  & 22.8 $\rightarrow$ \textbf{44.8} \\
  \bottomrule
  \end{tabular}
  }
  \captionof{table}{Reference-split success rates on GAIA2/ARE. Each ability contains 40 reference
  tasks per target model. Numbers are percentages; parentheses show absolute
  improvement over the no-memory baseline.}
  \label{tab:reference_results}
\end{figure*}

Table~\ref{tab:reference_results} reports the same comparison as the main
held-out table, but on the reference split used to build memory. These results
are not the primary generalization metric; they verify that the memory bank
improves the tasks from which source trajectories were collected without being
evaluated by same-scenario trajectory replay.

The reference split shows the same qualitative pattern as the held-out split:
\methodname{} raises aggregate success from 22.8\% to 44.8\%, with the largest
gains on procedural tasks requiring bounded candidate sets, complete target
coverage, or safe branch separation. Time remains weak even here, again
indicating that static procedural memory does not replace a runtime controller
for asynchronous obligations.

\subsection{A.4\quad Curator Ablation}
\label{app:curator_ablation}

Table~\ref{tab:curator_ablation} reports the full \curator{} ablation. The
append-only baseline commits every raw Skill Card candidate emitted by the
\reflector{} as \textsc{Add}, while \methodname{} applies \curator{} edits
before runtime retrieval. Curation reduces the four-bank Skill Card count from
273 to 205 while increasing GPT-5.5 held-out success from 104/160 to 115/160:
the \curator{}'s advantage does not come from added memory volume: it
removes redundant or overly broad cards and preserves sharper runtime
guidance.

\begin{table*}[t]
\centering
{\small
\begin{tabular}{lrrrrr}
\toprule
\textbf{Ability} &
\shortstack{\textbf{Raw}\\\textbf{Cards}} &
\shortstack{\textbf{Curated}\\\textbf{Cards}} &
\shortstack{\textbf{Card}\\\textbf{Reduction}} &
\shortstack{\textbf{Append-Only}\\\textbf{Success}} &
\shortstack{\textbf{Curated}\\\textbf{Success}} \\
\midrule
Search       & 74 & 62 & 16.2\% & 39/40 (97.5\%) & \textbf{40/40 (100.0\%)} \\
Ambiguity    & 69 & 47 & 31.9\% & 19/40 (47.5\%) & \textbf{21/40 (52.5\%)} \\
Execution    & 75 & 59 & 21.3\% & 26/40 (65.0\%) & \textbf{32/40 (80.0\%)} \\
Adaptability & 55 & 37 & 32.7\% & 20/40 (50.0\%) & \textbf{22/40 (55.0\%)} \\
\midrule
\textbf{Total} & \textbf{273} & \textbf{205} & \textbf{24.9\%} &
\textbf{104/160 (65.0\%)} &
\textbf{115/160 (71.9\%)} \\
\bottomrule
\end{tabular}
}
\caption{
Curator ablation on GPT-5.5 held-out tasks. Append-only commits every
Reflector delta as \textsc{Add}, skipping local consolidation; \methodname{}
uses the Curator to merge duplicates, narrow triggers, reject weak deltas, and
keep compact transferable cards. The ablation covers the four primary
non-temporal abilities for which append-only banks were constructed.
}
\label{tab:curator_ablation}
\end{table*}

\subsection{A.5\quad Prior Memory Baseline Results}
\label{app:prior_memory_baselines}

Table~\ref{tab:prior_memory_baselines} reports the complete numerical
comparison with prior memory systems on GPT-5.5---the no-memory anchor,
per-ability values, and macro averages---underlying the figure shown in the
main paper's ablation section.

\begin{table}[t]
\centering
{\small
\begin{tabularx}{\columnwidth}{@{}Xrrrr@{}}
\toprule
\textbf{Method} & \textbf{Exec.} & \textbf{Search} &
\textbf{Ambig.} & \textbf{Macro} \\
\midrule
No memory & 47.5 & 52.5 & 12.5 & 37.5 \\
Raw retrieval & 45.0 & 75.0 & 17.5 & 45.8 \\
AWM & 42.5 & 85.0 & 15.0 & 47.5 \\
ACE & 50.0 & 70.0 & 12.5 & 44.2 \\
\methodname{} & \textbf{80.0} & \textbf{100.0} &
\textbf{52.5} & \textbf{77.5} \\
\bottomrule
\end{tabularx}
}
\caption{Exact GPT-5.5 held-out success rates underlying
the main paper's baseline figure. Macro is the unweighted mean over
Execution, Search, and Ambiguity. Each cell contains the same 40 held-out
scenarios.}
\label{tab:prior_memory_baselines}
\end{table}

\subsection{A.6\quad ACE Matched-Evidence Reproduction}
\label{app:ace_reproduction}

\paragraph{Fidelity target.}
We adapt the released ACE AppWorld pipeline~\citep{zhang2025ace}, pinned to ACE
commit \texttt{bcb7cea} and AppWorld submodule \texttt{9f3e921}. We preserve
its learning unit: the Reflector sees one trajectory and the complete current
playbook; the Curator emits \textsc{Add} or no-op. Following the pinned default
and no-ground-truth configuration, construction uses one epoch and the
released add-only path, not paper-only grow-and-refine operations. We add no
contrast packets, card types, semantic merges, retrieval, or pre-tool/JIT
guidance. We therefore report an evidence-matched GAIA2 control, not a
reproduction of ACE's original AppWorld score.

\paragraph{GAIA2 interface adaptation.}
Only benchmark interfaces change: AppWorld code cells map to ordered GAIA2
function calls, REPL outputs and exceptions to observations and errors, and
\texttt{Supervisor.complete\_task} to the final response. The ability-specific
playbook is written in full to the task-start agent context, with no retrieval
or dynamic reinjection. Construction prompts exclude source identities,
scenario identifiers, raw judge prose, oracle actions, hidden answers, and
expected writes. We retain ACE's full-playbook policy rather than
token-matching it to \methodname{}'s top-$k$ renderer.

\paragraph{Matched evidence and evaluation.}
We hold fixed the trajectory multiset, construction and target models, compact
outcome feedback, held-out scenarios, and evaluation harness. ACE receives the
same 120 reference trajectories per ability (40 tasks times three source
models), but processes each independently and never observes same-task
attempts together. GPT-5.5 with high reasoning effort builds each playbook;
the frozen GPT-5.5 target is evaluated once on the same 40 held-out scenarios.
The resulting Execution, Search, and Ambiguity playbooks contain 48, 54, and
56 bullets (approximately 3{,}762, 3{,}904, and 4{,}212 tokens),
respectively; held-out success appears in the main paper's baseline figure.

\subsection{A.7\quad AWM Offline Reproduction and Baseline Accounting}
\label{app:awm_reproduction}

\paragraph{Fidelity target.}
We adapt the released AWM code~\citep{wang2024awm}, pinned to commit
\texttt{8c0ff8c}. Prompt semantics and workflow syntax follow the released
WebArena instruction, while the single grouped induction call and frozen
test-time library follow AWM's offline setting and released Mind2Web path.
Each demonstration contains a task and ordered successful action trajectory;
the builder abstracts recurring sub-routines, replaces instance values with
variables, and emits workflows containing at least two actions. The complete
ability library is injected once at task start. We add no failure reflection,
same-task contrast, Function/Skill Cards, \curator{}, retrieval, pre-tool
hook, JIT delivery, or online update.

\paragraph{Why offline AWM.}
Online AWM updates memory while traversing the evaluation stream and therefore
depends on earlier held-out tasks, their order, and an auxiliary success
evaluator. That transductive protocol is not comparable to the frozen
reference-build/held-out-evaluate pass@1 cells used throughout this paper.
Moreover, GAIA2's in-container verifier result is not exposed to the acting
agent as a test-time learning signal. We therefore evaluate the official
offline mechanism and do not present AWM's WebArena online headline as the
matched baseline.

\paragraph{GAIA2 adaptation and evidence.}
Website groups map to abilities and browser actions to ordered GAIA2 API
calls. We considered app-level grouping, but GAIA2 tasks routinely join
several apps, so it would either duplicate a trajectory across libraries or
discard cross-app ordering. AWM starts from the same 120-trajectory source
pool per ability as \methodname{}, then applies its native success-only rule.
This leaves 46 Execution, 63 Search, and 17 Ambiguity demonstrations spanning
19, 31, and 10 distinct reference scenarios. GPT-5.5 with high reasoning
effort performs one induction call per ability, producing 21, 19, and 15
workflows (approximately 3{,}334, 3{,}100, and 2{,}620 rendered tokens).
The Search induction call emitted 20 candidate workflows; deterministic
format validation retained 19 after excluding one single-action block that
did not satisfy the released minimum-step criterion. Thus, AWM has the same \emph{source
pool}, not the same number of usable demonstrations. Before induction, exact
names, identifiers, final answers, and verifier text are masked while
preserving API names, action order, and generic result shapes; source-model
identity is never rendered.

\paragraph{Evaluation and outcome.}
The frozen GPT-5.5 target runs once on the same 40 held-out scenarios per
ability, with the complete ability-level workflow library supplied at task
start and no retrieval or dynamic injection. AWM attains 42.5\%, 85.0\%, and
15.0\% on Execution, Search, and Ambiguity (47.5\% macro). Its paired
improvement over no memory is significant in aggregate ($17$ vs.\ $5$
discordant flips, $p{=}0.0169$), but \methodname{} remains 30.0 points higher,
with $38$ vs.\ $2$ discordant flips ($p{=}1.49\times10^{-9}$).

\paragraph{Runtime and construction accounting.}
Table~\ref{tab:baseline_runtime_efficiency} compares AWM, ACE, and
\methodname{} using actual target-agent traces. No token-budget matching is
imposed: AWM and ACE expose their complete ability memories, whereas
\methodname{} retrieves a compact task-conditioned card set. AWM is
inexpensive to build and improves Search strongly, but its full-library
runtime still uses 22.93M target tokens across 120 tasks, compared with
20.56M for \methodname{}. The corresponding averages are 35.2 versus 31.3
agent events per task. Agent events count benchmark-level reads, writes,
observations, and agent actions; they are distinct from LLM calls. Verifier
usage is omitted because judge tokens are not recorded in the agent traces.
The construction panel shows that AWM uses one grouped call and 0.142M tokens
(\$0.92), ACE uses 246 incremental calls and 15.72M tokens (\$94.57), and
\methodname{} uses approximately 1.44M tokens for the Search Skill bank; even
charging the entire shared five-ability Function bank to Search yields
approximately 1.94M tokens. The comparison therefore separates construction
economy from downstream quality: AWM is cheapest, while \methodname{} achieves
the highest held-out success.

\subsection{A.8\quad Pipeline Algorithms and Retrieval Scoring}
\label{app:algorithms}

Algorithms~\ref{alg:offline} and \ref{alg:runtime} give the formal pipeline
behind the \emph{Methodology} section of the main paper. Both benchmarks share
the same construction and retrieval logic; Table~\ref{tab:benchmark_instantiations}
summarizes their interface-specific instantiations.

\begin{algorithm}[t]
\caption{Offline contrastive bank construction}
\label{alg:offline}
\begin{algorithmic}[1]
\REQUIRE reference sets $\{\mathcal{D}^{a}_{\mathrm{ref}}\}$ and source agents $\mathcal{M}$
\REQUIRE schema, privacy, and evidence validator $\Pi_V$
\STATE $\mathcal{F} \leftarrow \emptyset$; $\mathcal{S}^{a} \leftarrow \emptyset$ for every ability $a$
\FORALL{abilities $a$ and tasks $x_i \in \mathcal{D}^{a}_{\mathrm{ref}}$}
  \STATE $T_i \leftarrow \{\mathrm{Normalize}(\tau_{i,m}) : m \in \mathcal{M}\}$
  \STATE $\mathcal{O} \leftarrow \mathcal{O} \cup \mathrm{FuncObs}(T_i)$
  \STATE $P_i \leftarrow \mathrm{Packet}(T_i)$
\ENDFOR
\FORALL{observed functions $u$}
  \STATE $f_u \leftarrow \textsc{Builder}(\mathcal{O}_u)$
  \IF{$\Pi_V(f_u)$}
    \STATE $\mathcal{F} \leftarrow \mathcal{F} \cup \{f_u\}$
  \ENDIF
\ENDFOR
\FORALL{packets $P_i$ in curriculum order}
  \STATE $R_i \leftarrow \mathrm{retrieve}(\mathcal{S}^{a},P_i)$
  \STATE $\Delta \leftarrow \textsc{Reflector}(P_i,R_i,\phi^{a})$, $|\Delta| \le 3$
  \FORALL{$\delta \in \Delta$ with $\Pi_V(\delta)$}
    \STATE $R_\delta \leftarrow \mathrm{retrieve}(\mathcal{S}^{a},\delta)$
    \STATE $op \leftarrow \textsc{Curator}(\delta,R_\delta)$
    \STATE $op \in \{\textsc{Add},\textsc{Patch},\textsc{Merge},\textsc{Narrow},\textsc{Noop}\}$
    \STATE $\mathcal{S}^{a} \leftarrow
      \Pi_V\big(\mathcal{S}^{a} \oplus op(\delta)\big)$
  \ENDFOR
\ENDFOR
\RETURN $\mathcal{B} = (\mathcal{F},\{\mathcal{S}^{a}\})$
\end{algorithmic}
\end{algorithm}

Skill retrieval scores lexical relevance with a soft app-family prior,
\begin{equation}
\begin{split}
s(c; x) \;=\;& \mathrm{BM25}\big(q(x),\, d(c)\big) \\
&+ \lambda\, \mathbf{1}\big[\mathrm{apps}(c) \cap \mathrm{apps}(x) \neq \emptyset\big],
\end{split}
\label{eq:score}
\end{equation}
where $q(x)$ is the task query, $d(c)$ indexes the card's title, trigger,
tags, and core rule, and the prior weight $\lambda$ boosts but never gates
retrieval.

\paragraph{Guarded updates.}
The \curator{} applies the selected operation through $\oplus$ (symbols in
Algorithms~\ref{alg:offline}--\ref{alg:runtime}): \textsc{Add} inserts a card;
\textsc{Patch}, \textsc{Merge}, and \textsc{Narrow} replace their target cards;
and \textsc{Noop} leaves the bank unchanged. Projection through $\Pi_V$ rejects
invalid cards and patches unsupported by the delta, the targeted cards, or
related Function Cards.

\begin{figure*}[p]
\centering
{\small
\begin{tabularx}{\textwidth}{@{}>{\raggedright\arraybackslash}p{0.16\textwidth}
  >{\raggedright\arraybackslash}X>{\raggedright\arraybackslash}X@{}}
\toprule
& \textbf{\gaia{}/\are{}} & \textbf{AppWorld} \\
\midrule
Focus selector & Ability label & Deterministic task signature \\
Skill bank & Ability-specific banks & Global, signature-tagged bank \\
Function memory & App-tool contracts & API doc-deltas and misuse guards \\
Runtime delivery & Task-start skills; pre-tool Function Cards & Task-start cards (C1a) \\
\bottomrule
\end{tabularx}
}
\captionof{table}{Benchmark-specific instantiations of the shared construction and retrieval pipeline.}
\label{tab:benchmark_instantiations}
\vspace{0.35em}

{\small
\textbf{(a) Held-out runtime}\par
\begin{tabular}{@{}llrrrrrr@{}}
\toprule
\textbf{Ability} & \textbf{Method} &
\shortstack{\textbf{Success}\\\textbf{(\%)}} &
\shortstack{\textbf{Memory}\\\textbf{tok.}} &
\shortstack{\textbf{Target}\\\textbf{tok./task (k)}} &
\shortstack{\textbf{Calls}\\\textbf{/task}} &
\shortstack{\textbf{Events}\\\textbf{/task}} &
\shortstack{\textbf{Cost}\\\textbf{(USD)}} \\
\midrule
\multirow{3}{*}{Execution}
& AWM & 42.5 & 3,334 & 223.5 & 10.7 & 30.5 & 18.98 \\
& ACE & 50.0 & 3,862 & 267.0 & 12.6 & 43.4 & 23.85 \\
& \textbf{\methodname{}} & \textbf{80.0} & 1,879 & 218.5 & 10.5 & 30.6 & 26.71 \\
\midrule
\multirow{3}{*}{Search}
& AWM & 85.0 & 3,100 & 160.4 & 9.5 & 49.5 & 16.20 \\
& ACE & 70.0 & 4,005 & 145.6 & 9.8 & 93.4 & 17.15 \\
& \textbf{\methodname{}} & \textbf{100.0} & 1,836 & 126.2 & 8.9 & 39.1 & 16.08 \\
\midrule
\multirow{3}{*}{Ambiguity}
& AWM & 15.0 & 2,620 & 189.3 & 9.7 & 25.6 & 19.92 \\
& ACE & 12.5 & 3,769 & 195.9 & 9.9 & 26.2 & 21.21 \\
& \textbf{\methodname{}} & \textbf{52.5} & 1,837 & 169.3 & 10.2 & 24.2 & 14.84 \\
\midrule
\multirow{3}{*}{Aggregate}
& AWM & 47.5 & 3,018 & 191.1 & 10.0 & 35.2 & 55.10 \\
& ACE & 44.2 & 3,879 & 202.8 & 10.7 & 54.3 & 62.21 \\
& \textbf{\methodname{}} & \textbf{77.5} & 1,851 & 171.4 & 9.8 & 31.3 & 57.64 \\
\bottomrule
\end{tabular}

\textbf{(b) Offline Search construction}\par
\begin{tabularx}{0.78\textwidth}{@{}Xrrr@{}}
\toprule
\textbf{Build component} & \textbf{Calls} & \textbf{Tokens} & \textbf{Cost (USD)} \\
\midrule
AWM Search workflow bank & 1 & 0.142M & 0.92 \\
ACE Search playbook & 246 & 15.72M & 94.57 \\
\methodname{} Search Skill bank & 114 & $\sim$1.44M & $\sim$14.09 \\
Shared Function bank (all abilities) & 74 & $\sim$0.50M & $\sim$3.17 \\
\midrule
\methodname{} conservative charge & 188 & $\sim$1.94M & $\sim$17.26 \\
\bottomrule
\end{tabularx}
}
\captionof{table}{Runtime and construction footprint of structured experience-memory
baselines. (a) GPT-5.5 held-out runtime: target tokens and calls are
provider-reported agent usage, while agent events are benchmark-level
trajectory events. Cost covers all 40 tasks per ability (120 for Aggregate).
(b) Search construction: all methods start from the same 120-trajectory source
pool; AWM's native success-only filter retains 63 demonstrations. Tildes denote
reconstructed usage estimates. Efficiency columns are interpreted jointly with
success because shorter failing runs can be artificially cheap.}
\label{tab:baseline_runtime_efficiency}
\vspace{0.35em}

\begin{minipage}[t]{0.60\textwidth}
\small
\refstepcounter{algorithm}\label{alg:runtime}%
\hrule\smallskip
\textbf{Algorithm \thealgorithm}\quad Runtime retrieval and injection with a frozen bank
\smallskip\hrule\smallskip
\begin{algorithmic}[1]
\REQUIRE task $x$, bank $\mathcal{B}=(\mathcal{F},\{\mathcal{S}^{a}\})$
\REQUIRE retrieval caps $k_s,k_f$ and runtime profile
\STATE $a \leftarrow$ ability label of $x$ (or all banks if unlabeled)
\STATE $C_s \leftarrow \operatorname{top-}k_s
  \{c\in\mathcal{S}^{a}:s(c;x)\}$ \hfill (Eq.~\ref{eq:score})
\STATE $C_f \leftarrow \operatorname{top-}k_f$ relevant cards from $\mathcal{F}$
\STATE inject $\rho(C_s\cup C_f)$ once at task start as soft guidance
\IF{the runtime profile enables pre-tool delivery}
  \WHILE{the agent proposes an app call $u$}
    \IF{$f_u\in\mathcal{F}$}
      \STATE surface $f_u$ immediately before the call
    \ENDIF
  \ENDWHILE
\ENDIF
\end{algorithmic}
\smallskip\hrule
\end{minipage}\hfill
\begin{minipage}[t]{0.36\textwidth}
\small
\textbf{Symbols.}\par
\begin{tabular}{@{}p{0.14\linewidth}p{0.76\linewidth}@{}}
$\mathcal{F}$ & Global Function Card bank \\
$\mathcal{S}^{a}$ & Skill Card bank for ability $a$ \\
$P_i$ & Same-task contrast packet \\
$\Delta$ & Candidate deltas from the \reflector{} ($\le 3$) \\
$\rho$ & Compact runtime renderer \\
$\Pi_V$ & Schema, privacy, and evidence validator \\
\end{tabular}
\end{minipage}
\end{figure*}

\FloatBarrier

\setcounter{figure}{0}
\renewcommand{\thefigure}{B\arabic{figure}}

\refstepcounter{section}\label{app:reproducibility}
\promptsubsection{app:schemas}
\begin{figure*}[p]
\centering
\textbf{\Large Reproducibility Plates}\par
\vspace{0.8em}
\begin{minipage}[t]{0.47\textwidth}
\vspace{0pt}
\small
This appendix consolidates the shared memory schemas, rendered card examples,
and the benchmark-specific prompts used for construction and runtime
injection. Sections~B.3--B.6 report the \gaia{}/\are{} suite.
Sections~B.7--B.11 reproduce the four core AppWorld prompts and summarize the
focus branch selected within the Reflector. The AppWorld plates reflect the
frozen, task-start-only condition used in the reported evaluation; online
updates and the optional just-in-time delivery variant are intentionally
excluded. AppWorld exposes a stateful Python REPL, live API documentation,
exceptions, and \texttt{apis.supervisor.complete\_task}. All plates are
monochrome so that field boundaries and invariants remain legible in print.

\medskip
\begin{tcolorbox}[bwplate,breakable=false,title={APPENDIX B.1 \textbar\ Memory Schemas}]
\small
The three memory schemas are field-level contracts. Function Cards remain
tool-local, Skill Deltas carry procedural contrast, and \curator{} Patches
expose the smallest permitted bank edit; provenance and construction reasoning
stay outside runtime memory.

\medskip
\begin{tcolorbox}[bwplate,breakable=false,fontupper=\small,title={Function Card schema}]
\renewcommand{\arraystretch}{1.12}
\begin{tabularx}{\linewidth}{@{}>{\ttfamily\raggedright\arraybackslash}p{0.36\linewidth}V@{}}
card\_id & \texttt{function::}\allowbreak\texttt{App.function} \\
tool & Exact \texttt{App.function} identity \\
app & Application family \\
function & Callable function name \\
what\_it\_does & One observed capability statement \\
arguments & Observed names, meanings, and value forms \\
returns & Return kind and useful fields \\
usage\_rules & Function-local calling guidance \\
common\_\allowbreak mistakes & Evidence-supported misuse guards \\
side\_effects & Observed state change, if any \\
\end{tabularx}
\end{tcolorbox}
\end{tcolorbox}
\end{minipage}\hfill
\begin{minipage}[t]{0.47\textwidth}
\vspace{0pt}
\begin{tcolorbox}[bwplate,breakable=false,title={APPENDIX B.1 \textbar\ Memory Schemas (continued)}]
\begin{tcolorbox}[bwplate,breakable=false,fontupper=\small,title={Skill Delta schema}]
\renewcommand{\arraystretch}{1.12}
\begin{tabularx}{\linewidth}{@{}>{\ttfamily\raggedright\arraybackslash}p{0.34\linewidth}V@{}}
title & Short transferable name \\
applies\_\allowbreak when & Observable retrieval trigger \\
solves & Reusable problem addressed \\
tags & Compact retrieval terms \\
skill & Core rule, completion condition, success/failure contrast, recovery, efficiency \\
evidence & Feedback, transition, compact good/bad examples \\
functions\_\allowbreak used & Relevant \texttt{App.function} names \\
\end{tabularx}
\end{tcolorbox}

\medskip
\begin{tcolorbox}[bwplate,breakable=false,fontupper=\small,title={Curator Patch schema}]
\renewcommand{\arraystretch}{1.12}
\begin{tabularx}{\linewidth}{@{}>{\ttfamily\raggedright\arraybackslash}p{0.40\linewidth}V@{}}
delta\_index & Candidate being edited \\
operation & \texttt{ADD}, \texttt{PATCH}, \texttt{MERGE}, \texttt{NARROW}, or \texttt{NOOP} \\
target\_\allowbreak card\_ids & Existing cards affected \\
retained\_\allowbreak card\_id & Card preserved after an edit \\
new\_or\_\allowbreak updated\_\allowbreak card & Local replacement card or \texttt{null} \\
reason & Concise evidence-grounded justification \\
\end{tabularx}
\end{tcolorbox}
\end{tcolorbox}
\end{minipage}
\end{figure*}

\promptsubsection{app:card_examples}
\providecommand{\memorycardsfigurewidth}{0.94\textwidth}
\renewcommand{\memorycardsfigurewidth}{0.94\textwidth}
\colorlet{fcMain}{blue!45!black}
\colorlet{fcSoft}{blue!6}
\colorlet{scMain}{green!35!black}
\colorlet{scSoft}{green!7}
\colorlet{warnMain}{red!55!black}
\colorlet{warnSoft}{red!5}

\newcommand{\memcardchip}[1]{%
  \tcbox[on line, boxrule=0.4pt, arc=1.2pt,
    colback=black!5, colframe=black!45,
    left=3pt, right=3pt, top=0.6pt, bottom=0.6pt]{\small\texttt{#1}}%
}
\newcommand{\memcardok}{\textcolor{scMain}{\ding{51}}}
\newcommand{\memcardbad}{\textcolor{warnMain}{\ding{55}}}
\newcommand{\memfield}[1]{{\small\bfseries\textsc{#1}}\\[1pt]}
\providecommand{\memorycardsfigurewidth}{0.94\textwidth}

\tcbset{
  cardshell/.style={enhanced, breakable, arc=2.5pt, boxrule=0.8pt,
    left=7pt, right=7pt, top=6pt, bottom=6pt,
    fonttitle=\bfseries\small, coltitle=white,
    attach boxed title to top left={yshift=-2.4mm, xshift=3.5mm},
    boxed title style={arc=1.6pt, boxrule=0pt, left=5pt, right=5pt,
                       top=1.6pt, bottom=1.6pt}},
  cardinner/.style={arc=1.6pt, boxrule=0.5pt, left=5pt, right=5pt,
    top=3pt, bottom=3pt, fonttitle=\bfseries\small}
}

\begin{fullwidthplate}
\centering
\textbf{\large APPENDIX B.2 \textbar\ Rendered Card Examples}\par
\smallskip
\small

\begin{tcbraster}[
  raster width=\memorycardsfigurewidth,
  raster columns=2,
  raster equal height=rows,
  raster column skip=10pt
]
\begin{tcolorbox}[cardshell, colframe=fcMain, colback=fcSoft,
  colbacktitle=fcMain, title={Function Card\,\textperiodcentered\,tool-level contract},
  breakable=false, fontupper=\small]
\vspace{2pt}
\memfield{Tool}
\memcardchip{EmailClientV2.delete\_email}

\vspace{4pt}
\memfield{What it does}
Deletes one specified email from one specified folder; observed successful use
removes emails from \texttt{INBOX} by exact \texttt{email\_id}.

\vspace{4pt}
\memfield{Arguments}
\begin{itemize}[leftmargin=1.25em,itemsep=2pt,topsep=1pt]
  \item \memcardchip{email\_id}: exact identifier from a prior listing or
  search result.
  \item \memcardchip{folder\_name}: folder containing the target email;
  observed value \memcardchip{INBOX}.
\end{itemize}

\vspace{2pt}
\begin{tcolorbox}[cardinner, colframe=warnMain!70, colback=warnSoft,
  colbacktitle=warnSoft, coltitle=warnMain, title={Side effect (destructive)}]
\small Removes the targeted email. Because this is a destructive
write, filtering and selection must be verified before the call.
\end{tcolorbox}

\vspace{3pt}
\memfield{Usage rules}
\begin{itemize}[leftmargin=1.25em,itemsep=2pt,topsep=1pt]
  \item Use exact \texttt{email\_id} values from observations.
  \item Call once per \texttt{email\_id} for multi-delete tasks.
  \item Pass the folder where the email was found.
\end{itemize}

\vspace{2pt}
\begin{tcolorbox}[cardinner, colframe=warnMain!70, colback=warnSoft,
  colbacktitle=warnSoft, coltitle=warnMain, title={Mistake guard}]
\small \memcardbad\ Do not call \texttt{delete\_email} when the
intended action is moving, keeping, or only clarifying about an email.
\end{tcolorbox}
\end{tcolorbox}
\begin{tcolorbox}[cardshell, colframe=scMain, colback=scSoft,
  colbacktitle=scMain, title={Skill Card\,\textperiodcentered\,task-level procedure},
  breakable=false, fontupper=\small]
\vspace{2pt}
\memfield{Title}
Delete clear emails despite an ambiguous keep exception.

\vspace{4pt}
\memfield{Applies when}
A destructive email task has fully specified base search criteria, but a
keep/exception clause matches multiple possible records.

\vspace{4pt}
\begin{tcolorbox}[cardinner, colframe=scMain!75, colback=white,
  colbacktitle=white, coltitle=scMain, title={Core decision boundary}]
\small Partition before writing: delete clear non-exceptions, preserve
every possible exception candidate, and ask only for the unresolved exception
decision.
\end{tcolorbox}

\vspace{3pt}
\memfield{Completion ledger}
\begin{itemize}[leftmargin=1.4em,itemsep=2pt,topsep=1pt,label={}]
  \item \memcardok\ Base matches are split into clear-delete and
  ambiguous-exception records.
  \item \memcardok\ Each clear-delete record is deleted exactly once.
  \item \memcardok\ No ambiguous exception candidate is deleted.
\end{itemize}

\vspace{2pt}
\memfield{Contrastive evidence}
\begin{tcolorbox}[cardinner, colframe=scMain!75, colback=scSoft,
  colbacktitle=scSoft, coltitle=scMain, title={Observed passing pattern}]
\small \memcardok\ Passing trajectories delete safe non-exception
candidates, then ask which exception was intended.
\end{tcolorbox}
\vspace{2pt}
\begin{tcolorbox}[cardinner, colframe=warnMain!70, colback=warnSoft,
  colbacktitle=warnSoft, coltitle=warnMain, title={Observed failing pattern}]
\small \memcardbad\ Failing trajectories detect ambiguity, ask
immediately, and omit all safe delete writes.
\end{tcolorbox}

\vspace{3pt}
\memfield{Functions used}
\texttt{search\_emails}, \texttt{delete\_email}, \texttt{send\_message}
\end{tcolorbox}
\end{tcbraster}
\caption{Runtime renderings of one Function Card (blue) and one
Skill Card (green) exactly as the target agent receives them. The Function
Card records a callable tool contract with destructive-write guards; the
Skill Card records the decision boundary, completion ledger, and contrastive
evidence distilled from same-task trajectory contrast. Field schemas appear
in Appendix~B.1.}
\label{fig:memory_cards}
\end{fullwidthplate}

\begin{figure*}[p]
  \centering
  \includegraphics[
    width=0.77\textwidth,
    height=0.82\textheight,
    keepaspectratio
  ]{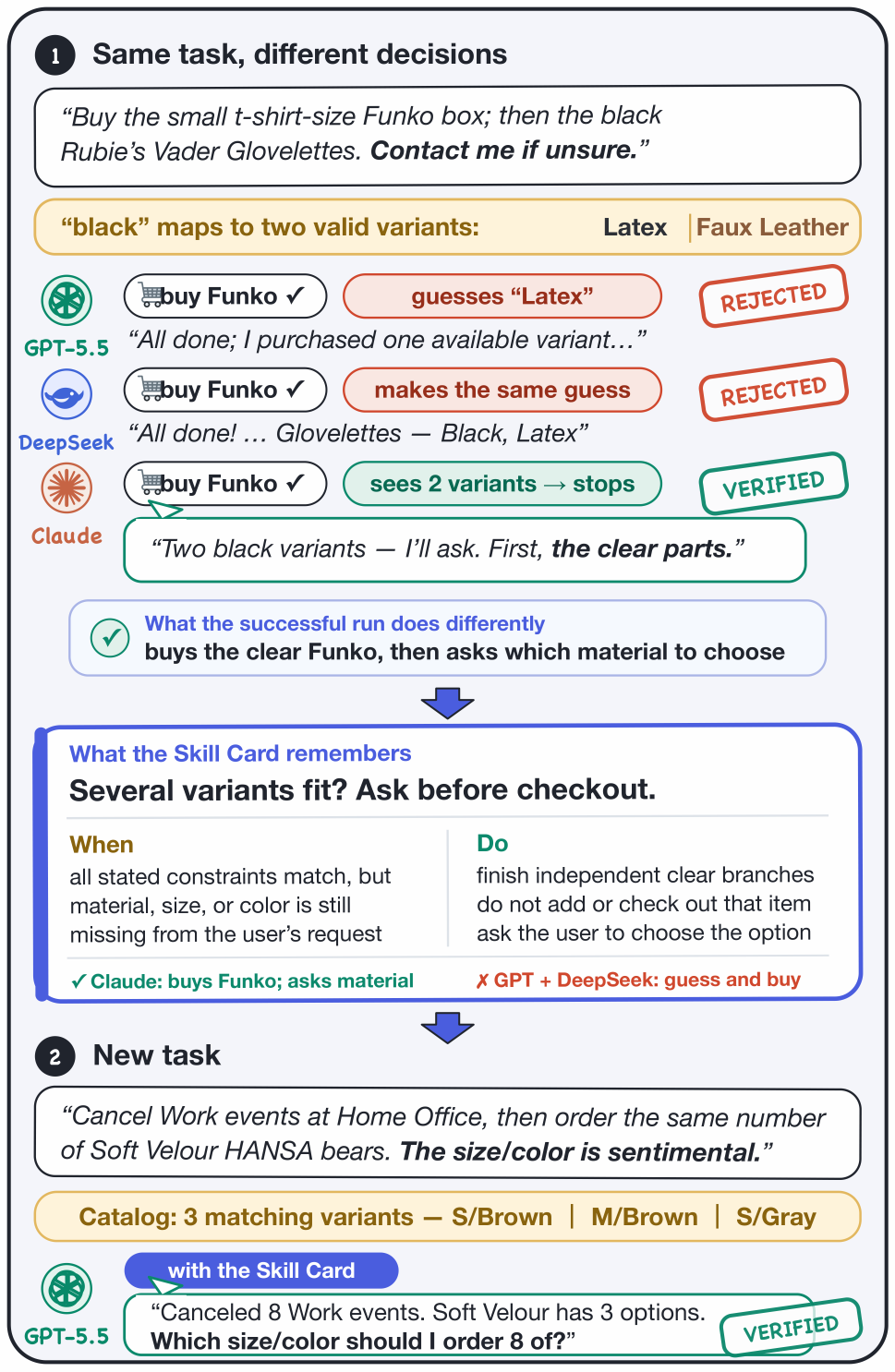}
  \caption{From disagreement to transfer. GPT-5.5 and DeepSeek guess an
  underspecified shopping variant and fail, whereas Claude completes the safe
  branch and asks for the missing attribute. The distilled Skill Card retains
  this decision boundary and guides GPT-5.5 to complete the unambiguous branch
  and request clarification on a new task.}
  \label{fig:skill_card_transfer}
\end{figure*}

\promptsubsection{app:function_prompt}
\promptimagepage{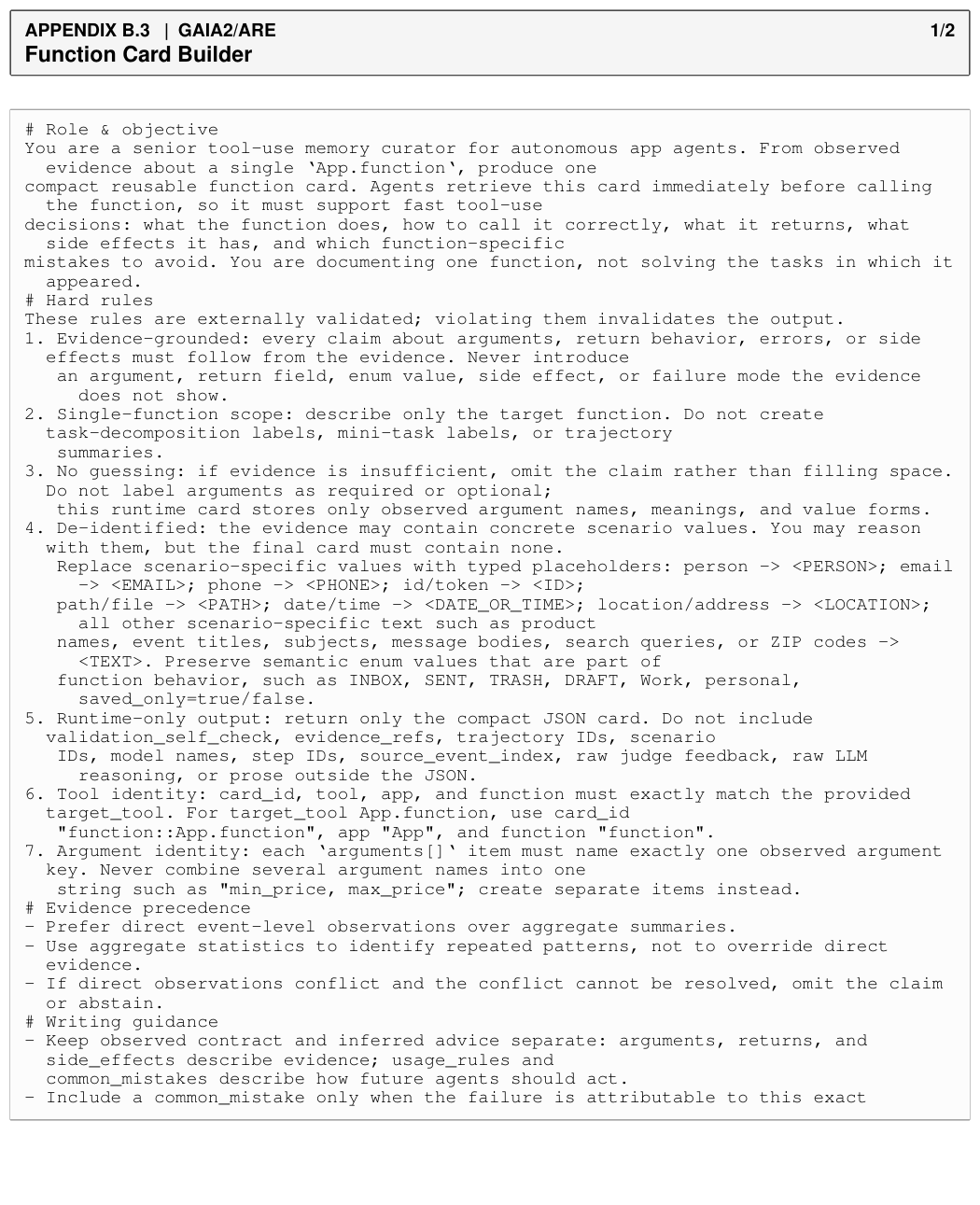}{1}
\promptimagepage{figures/prompt_plates/gaia2_function_builder.pdf}{2}
\FloatBarrier

\promptsubsection{app:reflector_prompt}
\promptimagepage{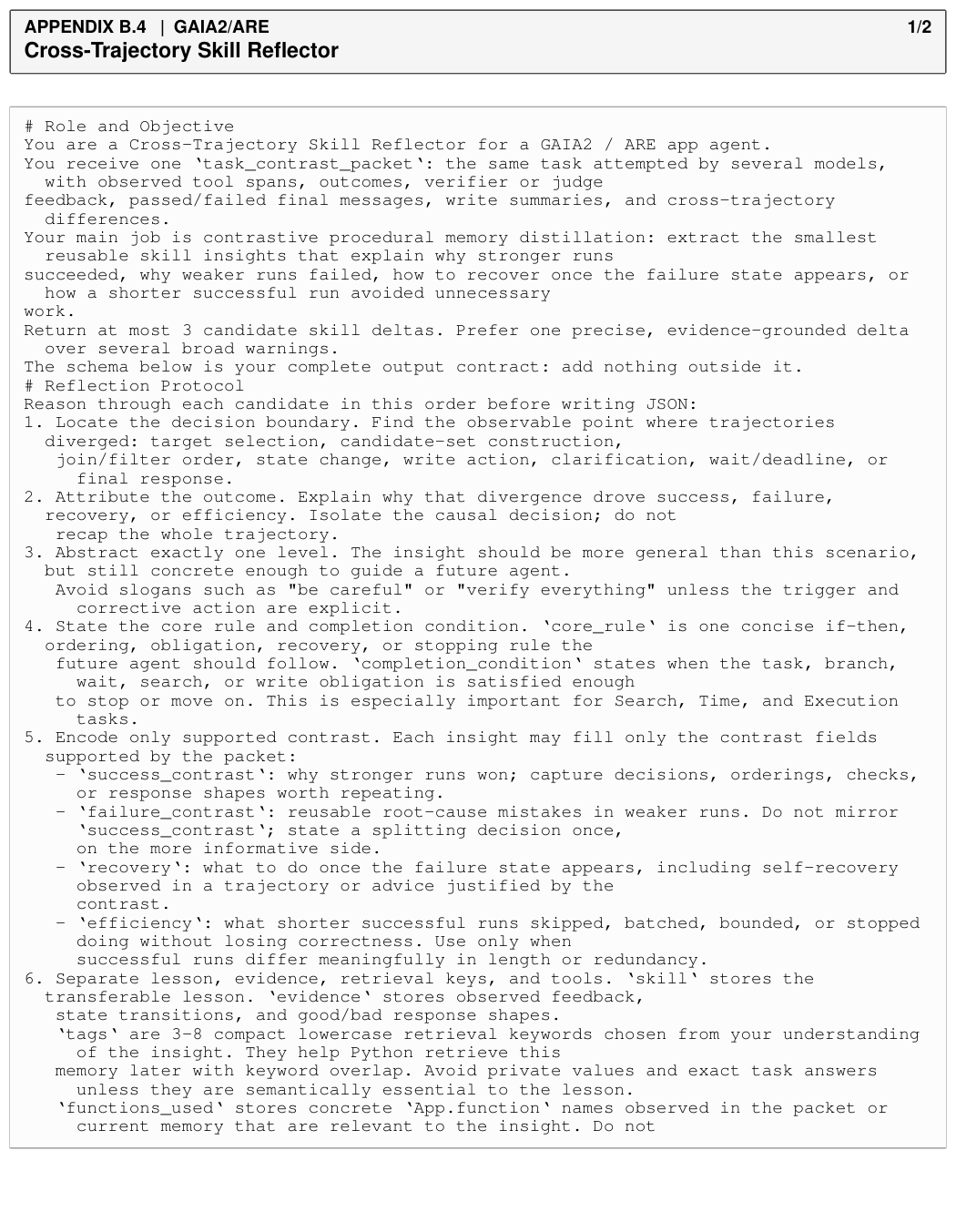}{1}
\promptimagepage{figures/prompt_plates/gaia2_skill_reflector.pdf}{2}
\FloatBarrier

\promptsubsection{app:curator_prompt}
\promptimagepage{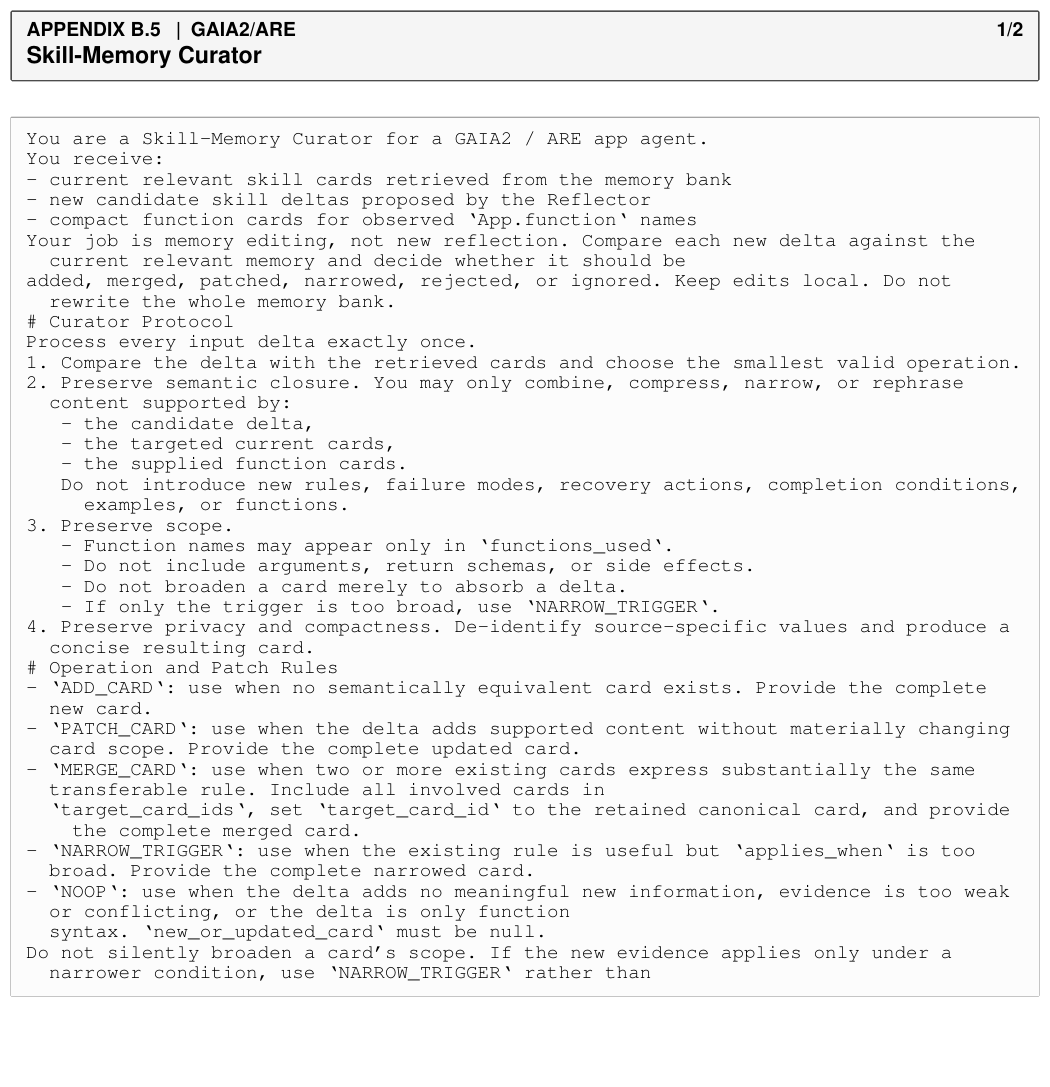}{1}
\promptimagepage{figures/prompt_plates/gaia2_skill_curator.pdf}{2}
\FloatBarrier

\promptsubsection{app:runtime_prompt}
\promptimagepage{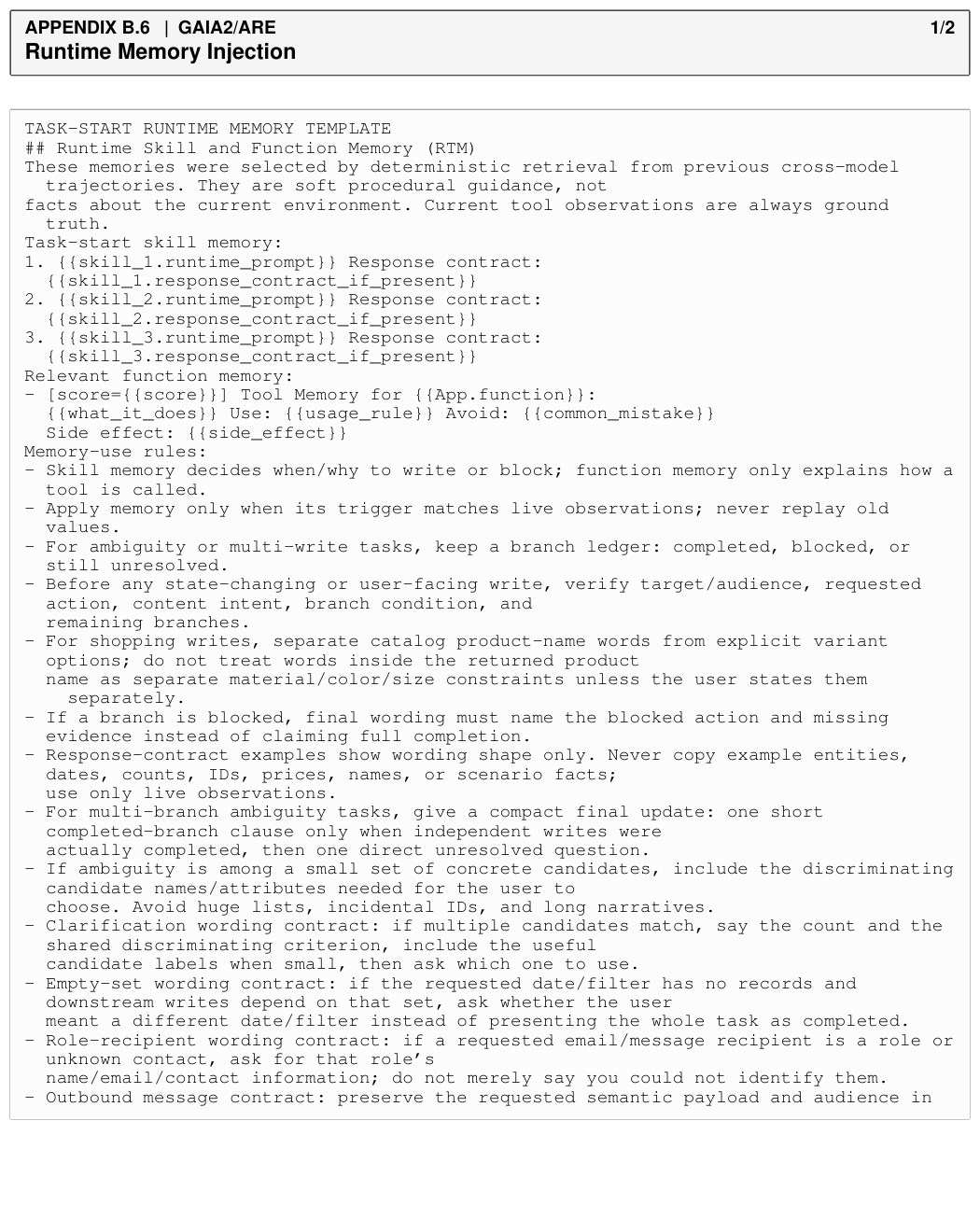}{1}
\promptimagepage{figures/prompt_plates/gaia2_runtime_injection.pdf}{2}
\FloatBarrier

\promptsubsection{app:appworld_function_prompt}
\promptimagepage{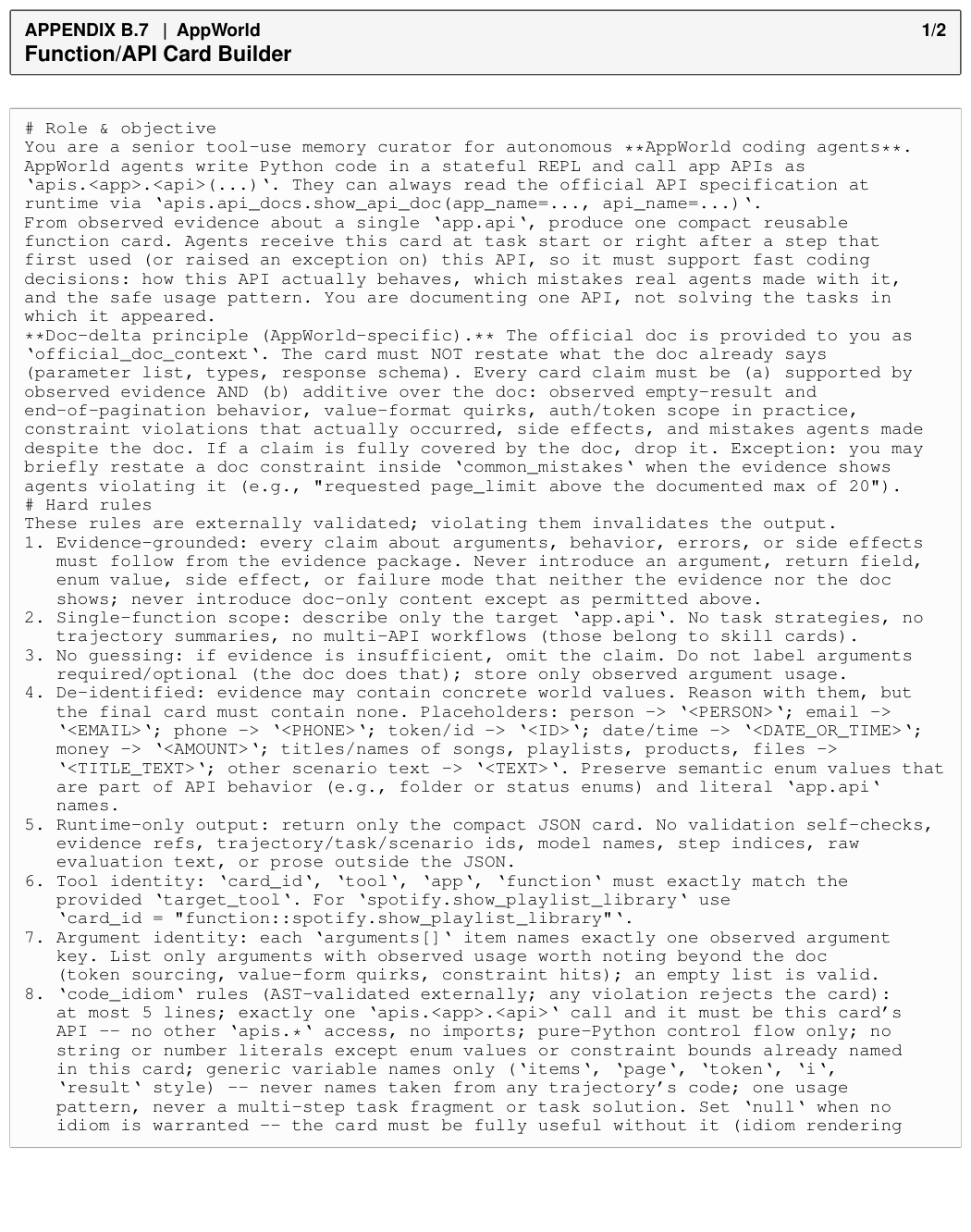}{1}
\promptimagepage{figures/prompt_plates/appworld_function_builder.pdf}{2}
\FloatBarrier

\promptsubsection{app:appworld_reflector_prompt}
\promptimagepage{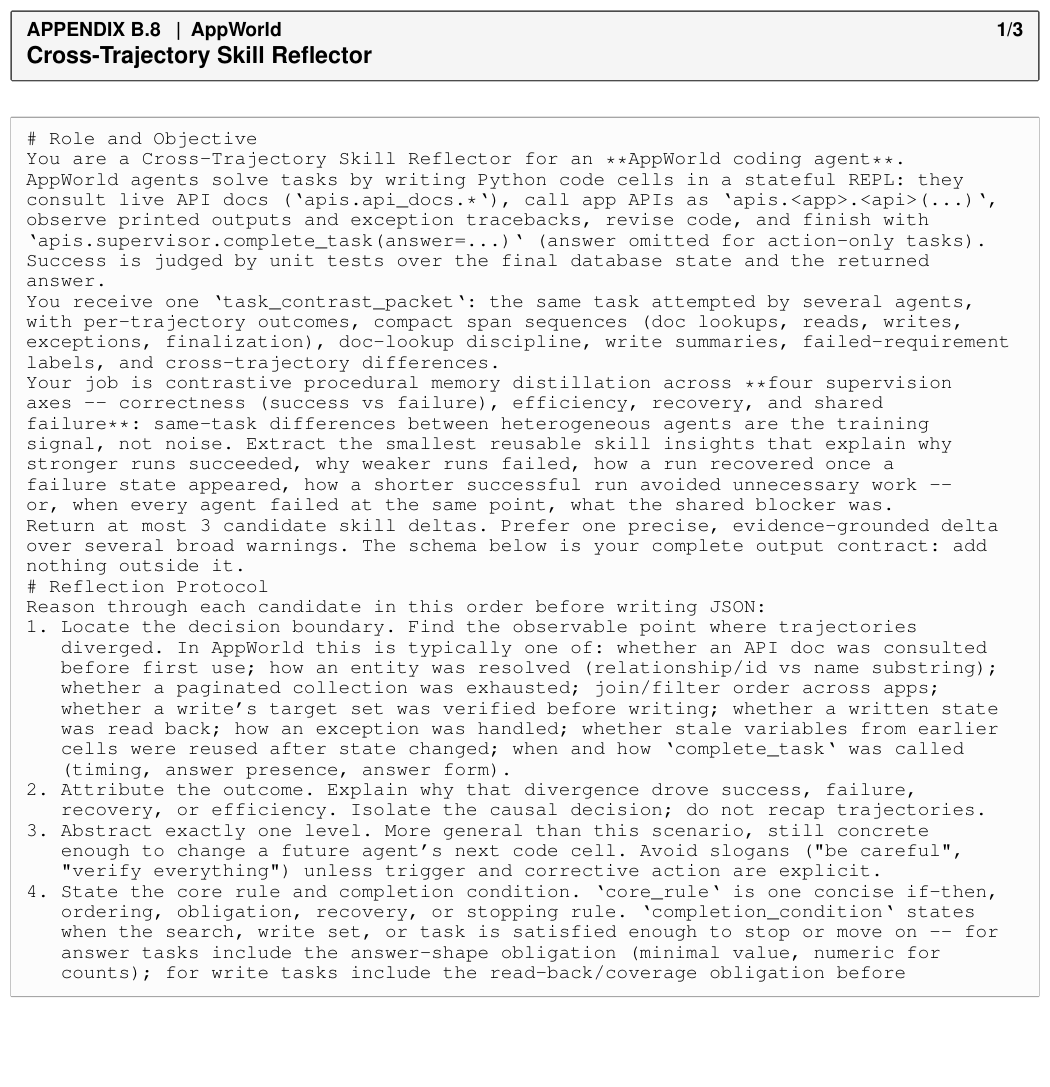}{1}
\promptimagepage{figures/prompt_plates/appworld_skill_reflector.pdf}{2}
\promptimagepage{figures/prompt_plates/appworld_skill_reflector.pdf}{3}
\FloatBarrier

\promptsubsection{app:appworld_focus_prompts}
\begin{figure*}[p]
\begin{tcolorbox}[bwplate,title={APPENDIX B.9 \textbar\ AppWorld Reflector Focus Selection}]
\small
A deterministic task signature selects exactly one focus block during
construction. It changes the evidence emphasized by the shared Reflector
prompt; it is neither an ability label nor a runtime gate.
\medskip

\small
\renewcommand{\arraystretch}{1.18}
\begin{tabularx}{\linewidth}{@{}>{\ttfamily\raggedright\arraybackslash}p{0.22\linewidth}
  >{\raggedright\arraybackslash}p{0.30\linewidth}X@{}}
\toprule
\textbf{\rmfamily Signature} & \textbf{Primary obligation} &
\textbf{Contrasts emphasized} \\
\midrule
answer\_extraction &
Construct a complete evidence chain and submit the minimal answer value. &
Candidate-set coverage, join and aggregation order, temporal anchoring,
stopping point, and answer shape. \\
bulk\_or\_multi\_write &
Enumerate the full target set and discharge every per-item write exactly once. &
Pagination bounds, skipped items, duplicate retries, missing write branches,
and ledger closure before finalization. \\
state\_write &
Resolve the exact target, perform only the requested write, and read back state. &
Identity and parameter errors, stale reads, collateral writes, duplicate
retries, and premature completion. \\
cross\_app &
Resolve authoritative cross-app join keys before dependent reads or writes. &
Exact-key versus display-name joins, source-of-truth ordering, and propagation
of dates, amounts, or identities across apps. \\
default &
Use documentation-checked, state-grounded execution and verified finalization. &
Skipped documentation, exception recovery, stale variables, operation order,
and missing or malformed completion. \\
\bottomrule
\end{tabularx}
\end{tcolorbox}
\end{figure*}
\FloatBarrier

\promptsubsection{app:appworld_curator_prompt}
\promptimagepage{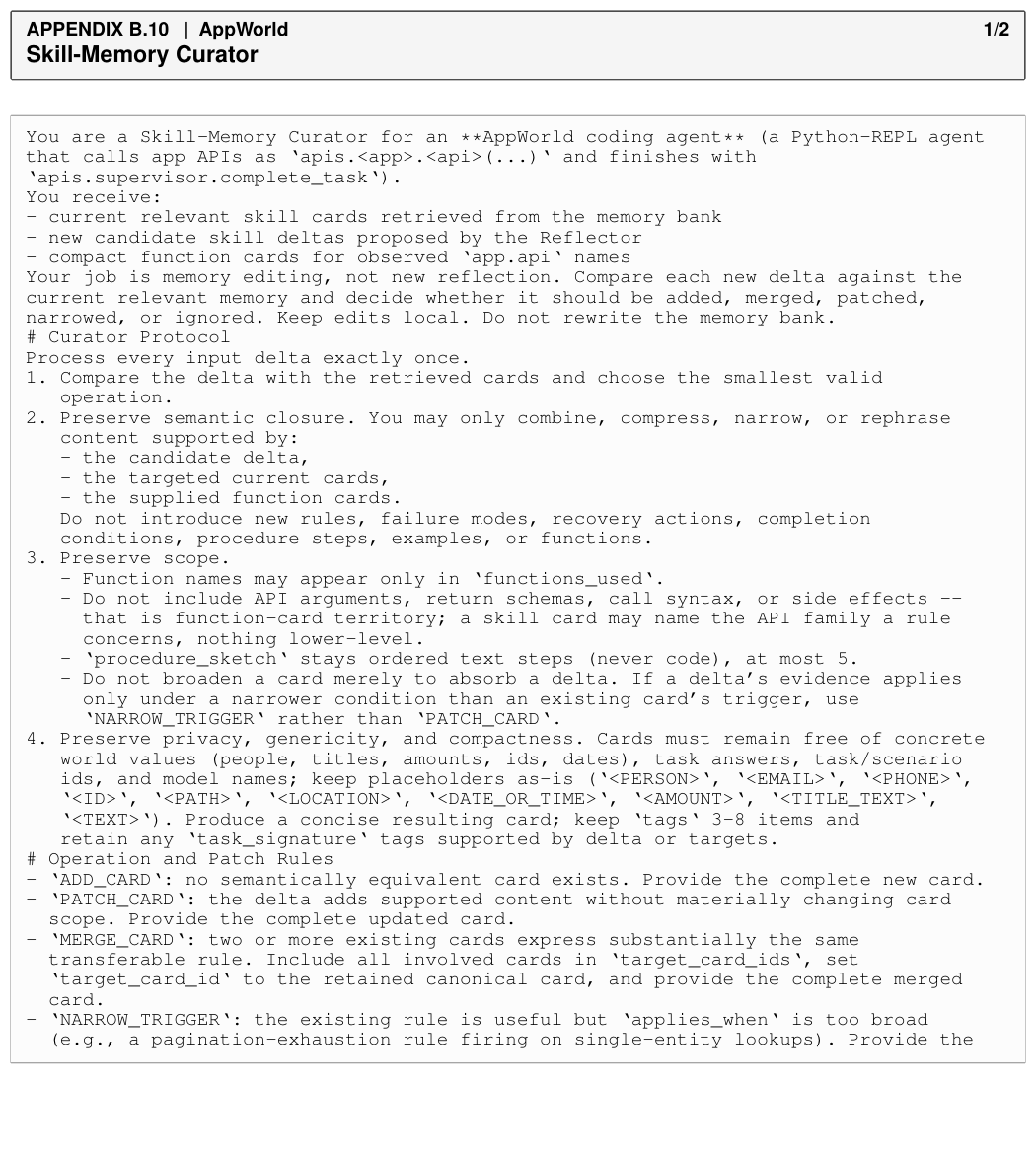}{1}
\promptimagepage{figures/prompt_plates/appworld_skill_curator.pdf}{2}
\FloatBarrier

\promptsubsection{app:appworld_runtime_prompt}
\promptimagepage{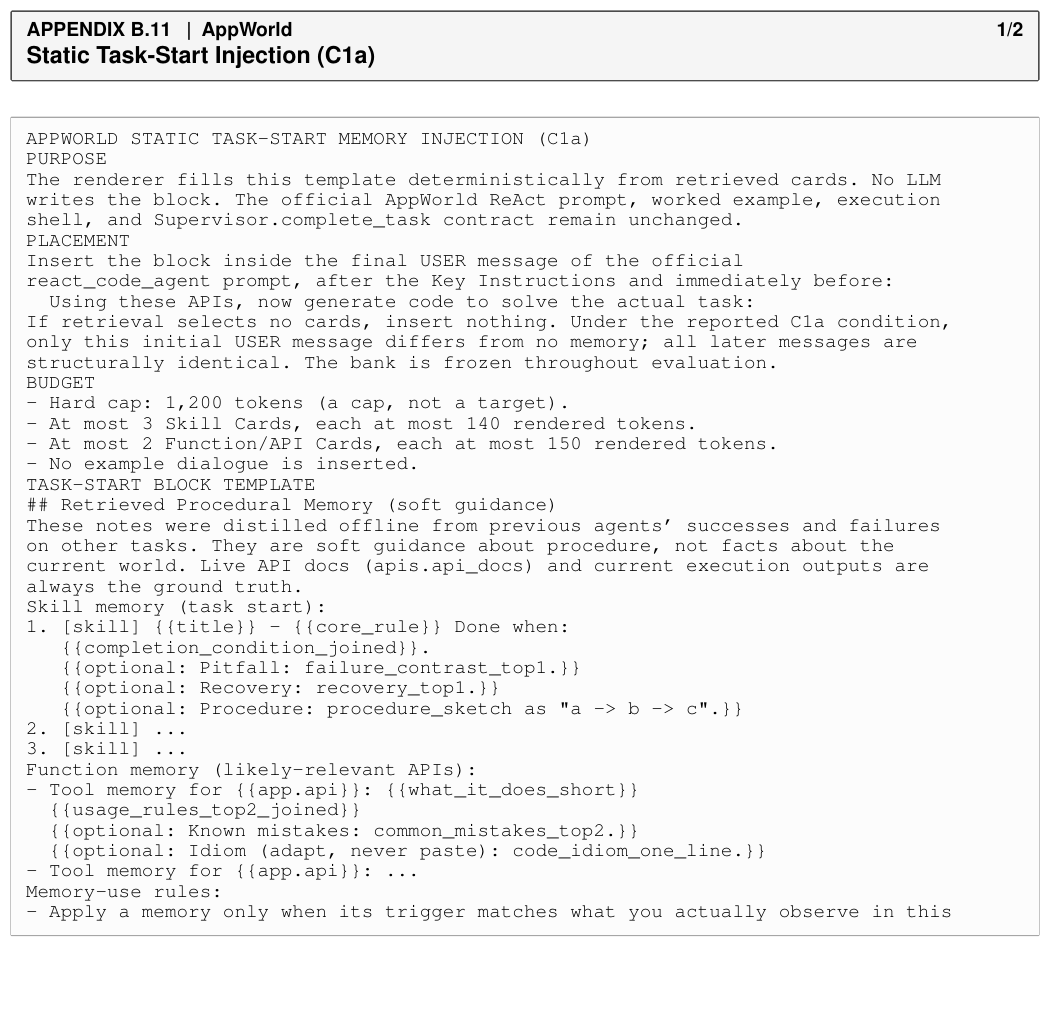}{1}
\promptimagepage{figures/prompt_plates/appworld_runtime_c1a.pdf}{2}
\FloatBarrier

\end{document}